\documentclass[10pt,twocolumn,a4paper]{article}
\usepackage[final]{cvww}

\usepackage{url}
\usepackage{booktabs}
\usepackage{algorithm}
\usepackage{algpseudocode}
\usepackage{graphicx}
\usepackage{xcolor}
\usepackage{pifont}
\usepackage{amsmath,amsfonts,bm}
\usepackage{subcaption}
\usepackage{adjustbox}

\definecolor{myred}{RGB}{187, 1, 54}
\definecolor{mygreen}{RGB}{61, 120, 56}
\definecolor{othergreen}{rgb}{0.0, 0.5, 0.0}
\definecolor{fgold}{RGB}{212, 175, 55}
\definecolor{ssilver}{RGB}{192, 192, 192}
\definecolor{tbronze}{RGB}{159, 122, 52}

\newcommand{\minisection}[1]{\vspace{0.04in} \noindent {\textbf{#1}\ }}

\newcommand{\cmark}{\textcolor{mygreen}{\ding{51}}}%
\newcommand{\xmark}{\textcolor{myred}{\ding{55}}}%
\newcommand{\spacerbullet}[1]{{#1}~\textcolor{white}{\scalebox{1.3}{$\bullet$}}}
\newcommand{\best}[1]{\textbf{#1}~\textcolor{fgold}{\scalebox{1.3}{$\bullet$}}}
\newcommand{\second}[1]{\textbf{#1}~\textcolor{ssilver}{\scalebox{1.3}{$\bullet$}}}
\newcommand{\third}[1]{\textbf{#1}~\textcolor{tbronze}{\scalebox{1.3}{$\bullet$}}}

\usepackage[pagebackref,breaklinks,colorlinks,allcolors=cvwwblue]{hyperref}

\def\paperID{9}

\title{Incremental Learning with Repetition via Pseudo-Feature Projection}

\author{
Benedikt Tscheschner$^{1,2}$\and
Eduardo Veas$^{1}$\and
Marc Masana$^{2,3}$\vspace{0.2em}\and
{$^{1}$Know-Center Research GmbH}\\
{$^{2}$Institute of Visual Computing, TU Graz}\\
{$^{3}$SAL Dependable Embedded Systems, Silicon Austria Labs}\\
{\tt\small \{btscheschner, eveas\}@know-center.at, mmasana@tugraz.at}
}

\begin{document}
\maketitle

\begin{abstract}
Incremental Learning scenarios do not always represent real-world inference use-cases, which tend to have less strict task boundaries, and exhibit repetition of common classes and concepts in their continual data stream. To better represent these use-cases, new scenarios with partial repetition and mixing of tasks are proposed, where the repetition patterns are innate to the scenario and unknown to the strategy. We investigate how exemplar-free incremental learning strategies are affected by data repetition, and we adapt a series of state-of-the-art approaches to analyse and fairly compare them under both settings. Further, we also propose a novel method (Horde), able to dynamically adjust an ensemble of self-reliant feature extractors, and align them by exploiting class repetition. Our proposed exemplar-free method achieves competitive results in the classic scenario without repetition, and state-of-the-art performance in the one with repetition.
\end{abstract}

\section{Introduction}
\label{sec:intro}
As autonomous agents and models in production systems are exposed to continuous streams of information, they are required to adapt to dynamic data distributions with potentially multiple tasks and integrate new information over time~\cite{verwimp2023continual, MASCHLER2021452, belouadah2020active}. The practice of retraining the complete system whenever new data is available becomes unfeasible as the storage, computation and privacy constraints for data streams increase~\cite{schwarz2018progress, parisi2019continual, radford2021learning}. To address these constraints, incremental learning~(IL) or continual learning has emerged as a promising approach~\cite{belouadah2021comprehensive}.

IL aims to learn a model sequentially through a sequence of tasks introducing disjoint sets of information at each training step~\cite{marc_survey2, marc_survey1, van2020brain}. Generally, these scenarios enforce a strict no-repetition constraint~\cite{incremental_repetition} allowing access to the data distribution only once in the task sequence. Unlike humans, who can learn nearly inference-free between tasks, neural networks suffer from a phenomenon called \emph{catastrophic forgetting}~\cite{empirical_catastrophic, french1999catastrophic}. When models are optimized sequentially on novel tasks, a swift forgetting of previously learned tasks is observed. To mitigate this forgetting, a delicate balance between preserving learned task knowledge (stability) and the ability to adapt to new information (plasticity) has to be reached, which is known as the stability-plasticity dilemma~\cite{stability_plasticity}. A popular approach to address this is to cache a representative subset of previously encountered data points in a buffer and replay them during the following training sessions~\cite{experience_based_il, van2020brain, rolnick2019experience}. Although such \emph{rehearsal} addresses catastrophic forgetting effectively, data privacy concerns have been raised~\cite{Li02012019}, and the scalability of an exemplar buffer in long-tailed incremental sequences is questionable~\cite{van2020brain} due to the large computational cost of complete retraining and significant storage requirements.

Nonetheless, the strict enforcement of no-class repetition becomes unrealistic for many real-world applications, as continuous streams are bound to repeat certain information~\cite{incremental_repetition} or be affected by semantic or covariate shifts~\cite{ovadia2019can}.
For example in industrial defect detection, certain common defects and defect-free samples will repeat throughout production.
The occurrence of repetition is further amplified in environments where an agent has the freedom to reexperience elements which are contained within the overall environment design.
Thus the effects of catastrophic forgetting are likely exaggerated as an uncontrollable form of rehearsal occurs naturally. Previous incremental learning research has largely explored catastrophic forgetting under the assumption that new information has a single opportunity to be learned, since each class is only available within a single task throughout the sequence. The introduction of repetition into these scenarios enables the selection of more broad incremental training tasks and highlights the different dynamics within the plasticity-stability dilemma of learning new tasks while maintaining current knowledge~\cite{incremental_repetition}.
The focus on catastrophic forgetting without repetition may limit the development of more realistic incremental learning agents, which involve different complex objectives like forward transfer~\cite{forward_transfer} and efficiency for computational limitations in edge devices~\cite{diaz2018don}.

\begin{figure}[t]
    \centering
    \vspace{1em}
    \includegraphics[width=0.95\linewidth]{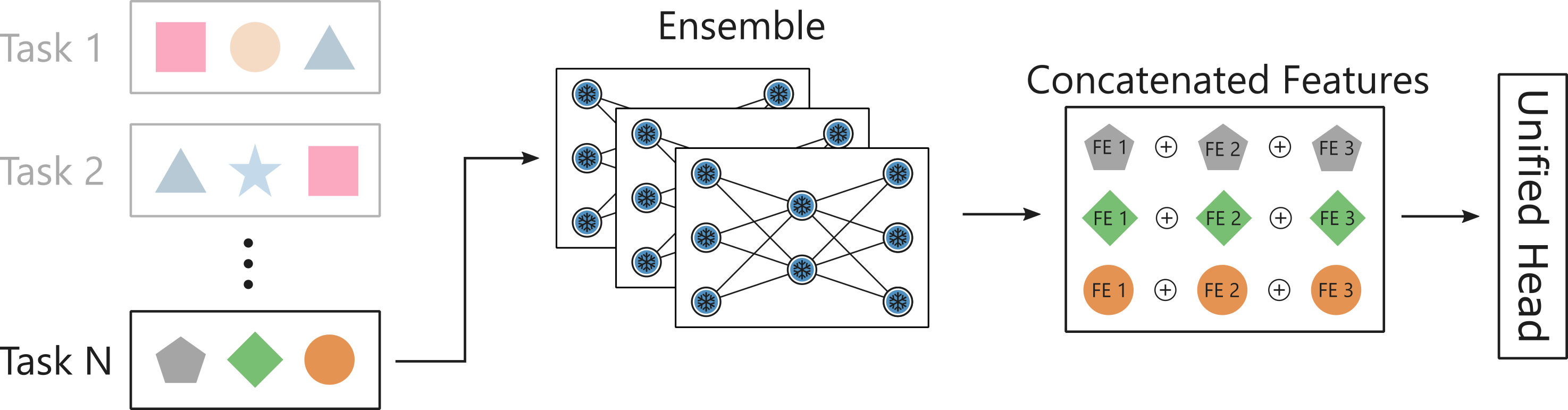}
    \caption{Overview of our proposed method (Horde). Each data sample is processed by an ensemble of independent feature extractors. The features from all extractors are concatenated before being passed into a unified head that can accommodate the dynamic input size through pseudo-feature projection.}
    \label{fig:horde_overview}
\end{figure}

As such, we want to loosen the no-repetition constraint and explore the effects of natural repetition. To explore these new settings and effects, our contributions are:
\begin{enumerate}
   \item a new variation of the class-incremental learning \mbox{CIFAR 50/10} scenario introducing class repetition,
   \item benchmarking a broad selection of state-of-the-art exemplar-free class-incremental learning methods and investigate the effects of innate data repetition and their resiliency to repetition frequency bias,
   \item a novel incremental learning method (Horde) that builds an ensemble of independent feature extractors for stability and utilizes pseudo-feature projection for plasticity (see \cref{fig:horde_overview}).
\end{enumerate}

\section{Related Work}
\label{sec:related}
Class-incremental learning (CIL) addresses the challenge of training a model sequentially on a series of tasks, without access to previous or future data~\cite{icarl}. When training without any constraints, models fail to retain knowledge from previous tasks -- a problem known as \emph{catastrophic forgetting}~\cite{french1999catastrophic, empirical_catastrophic}. Usually, each incremental task contains a disjoint set of new classes, which increases the difficulty of discriminating between those which have not been learned together under the same task~\cite{marc_survey2}.
A key challenge in incremental learning lies in keeping the balance of the stability-plasticity dilemma~\cite{stability_plasticity}, critical for mitigating catastrophic forgetting while ensuring the adaptability of the model to new tasks.

Incremental learning approaches include: weight regularization~\cite{ewc, mas}, which preserves important weights by estimating their importance; knowledge distillation~\cite{lwf, hinton2015distilling}, which focuses on protecting task representations rather than weights; rehearsal~\cite{rolnick2019experience}, which replay stored exemplars from previous tasks; mask-based approaches~\cite{masana2020ternary}, which use task-specific masks to isolate parameters that can be updated; and dynamic network structures~\cite{plastil, ensembles}, which expand the model architecture by adding new or contracting existing modules for each task. In this work, we concentrate on weight-regularization, knowledge distillation and dynamic network structure-based methods. These are the approaches that work on task-agnostic scenarios (do not require a task-ID during inference) and promote privacy preservation (do not store samples).

\minisection{Incremental learning with repetition.}
In many practical applications (automated failure inspection, medical imaging, robotics), pattern repetition naturally arises, yet traditional CIL approaches assume that each class is encountered only once, imposing a strict no-repetition constraint~\cite{incremental_repetition}. This constraint focuses on the prevention of catastrophic forgetting but also diverges from real-world scenarios where classes may reappear or shift over time. To address this, Hemati et. al~\cite{hemati2023class, HEMATI2025106920} propose an extension to the class-incremental learning scenario which models the repetition of individual classes outside of a single task. Unlike joint incremental or rehearsal-based learning, this repetition is innate to the learning scenario and cannot be adjusted. This emphasizes an experience-based scenario~\cite{experience_based_il}, which favours shorter training tasks that can sometimes only cover a part of the class distribution. Moreover, covering scenarios that lie between the classic offline incremental and the online ones.

Class-incremental learning with repetition has received increased interest in the research community, being a central element in the challenge tracks of the last two CLVISION challenge tracks at CVPR 2023 and 2024~\cite{HEMATI2025106920, clvision_challenge_2024}. In the 2023 edition, we competed with a base variant of our proposed method, although without elements for controlling ensemble growth (see \cref{sec:ensemble_fe}), self-supervision (see \cref{sec:abl_fe_training}) or applicability to variable network architectures.

\minisection{Class prototypes and pseudo-features.}
To enforce stability and alleviate class-recency biases in the classifier~\cite{marc_survey1}, Exemplar-Free Class Incremental \mbox{Learning~(EFCIL)} methods~\cite{il2a, pass, praka, ssre, fetril} utilize class prototypes to simulate unavailable classes. These prototypes capture statistical properties of embedding representations of each class, which are usually modeled as a multivariate Gaussian distribution~\cite{pass, il2a, praka}. Specifically, the statistics typically include the mean and covariance of feature representations for each class, allowing to generate pseudo-features when class data is not available. To extract representations, the neural network is divided into two modules. A feature extractor~(FE) that projects the input samples into their corresponding embedding representation; and a classifier head that uses these embeddings to solve the classification task.
Therefore, prototype-based methods can generate embeddings even when no samples from past classes are available during subsequent tasks by sampling the stored distributions of each class.
The sampled embedding representations are rehearsed alongside the current task data, thus promoting stability and mitigating class-recency bias. However, in order to maintain valid approximations of class distributions, the feature extractor needs to be either frozen or heavily regularized to prevent changes or drifts in the extracted features. Unlike rehearsal-based approaches, the use of prototypes does not violate data privacy due to the non-linearly projected representation in the embedding space~\cite{survey_class_incremental, pass}.
 
\minisection{Feature translation.} 
Instead of sampling the distribution approximated by class prototypes, FeTrIL~\cite{fetril} proposes to translate the features of available data classes to unavailable ones directly. Given a feature extractor $\mathbf{f}(x; \theta)$ being trained on current data $\{(x_{i},y_{i})\}$, its output embedding $F$ is efficiently translated from one of the current classes to the desired previously learned class $c\in \mathcal{Y}$ as
\begin{equation}
\hat{F}_{c} = \mathbf{f}(\bm{x}_{i}; \theta) + \mu_{c} - \mu_{\bm{y}_{i}}\ ,
\label{eq:fetril-proj}
\end{equation}

\noindent where $\mu_{c}$ and $\mu_{\bm{y}_{i}}$ represent the means of the old and current classes, respectively. The feature translation modifies the classifier, however, the feature extractor is requires to be frozen after the initial training so that the class means can be reliably extracted. This limits the continual learning process as the initial task constrains the diversity and robustness of the features that can be learned for new classes~\cite{belouadah2021comprehensive, fetril}. In our proposed approach, we relax this restriction by allowing an ensemble of smaller feature extractors to be learned. This allows for unknown class prototypes to be estimated through pseudo-feature projection until the repetition of classes allows for an accurate extraction of class prototypes.

\section{Method}
In incremental learning scenarios with repetition, the reappearance of classes introduces uncertainty in task sequences, requiring strategies that handle dynamic class distributions. Our approach aims to: (a) capture information from the current task, (b) integrate it with knowledge from previously seen tasks, and (c) ensure the ability to discriminate between all encountered classes so far. To achieve this, we leverage zero-forgetting feature extractors (FEs), which are aggregated in an ensemble to overcome the limitation of a completely fixed feature space. Through this aggregation, we form a flexible feature representation space that can adapt (expand or contract) based on the incremental learning sequence (see \cref{fig:horde_overview} for an overview of the proposed method structure).

\begin{figure}
    \centering
    \begin{subfigure}[c]{0.95\linewidth}
        \includegraphics[width=1.0\linewidth]{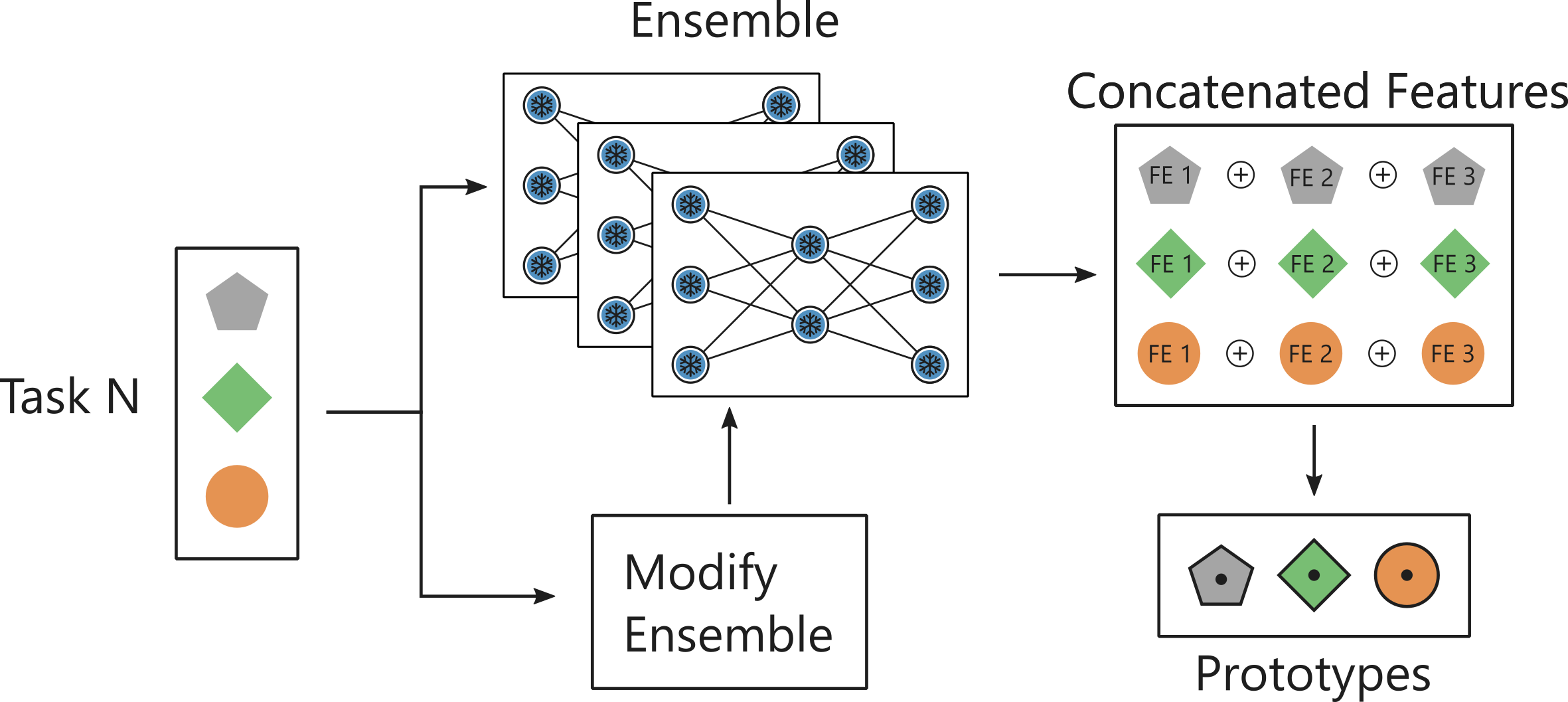}
        \caption{\textbf{Step 1}: The ensemble of feature extractors is adjusted based on the current task through either the addition or update of a self-reliant feature extractor. This step is only performed when estimated as necessary via a heuristic criteria.}
        \label{fig:step1_horde}
    \end{subfigure}
    \par\bigskip
    \begin{subfigure}[c]{0.95\linewidth}
        \includegraphics[width=1.0\linewidth]{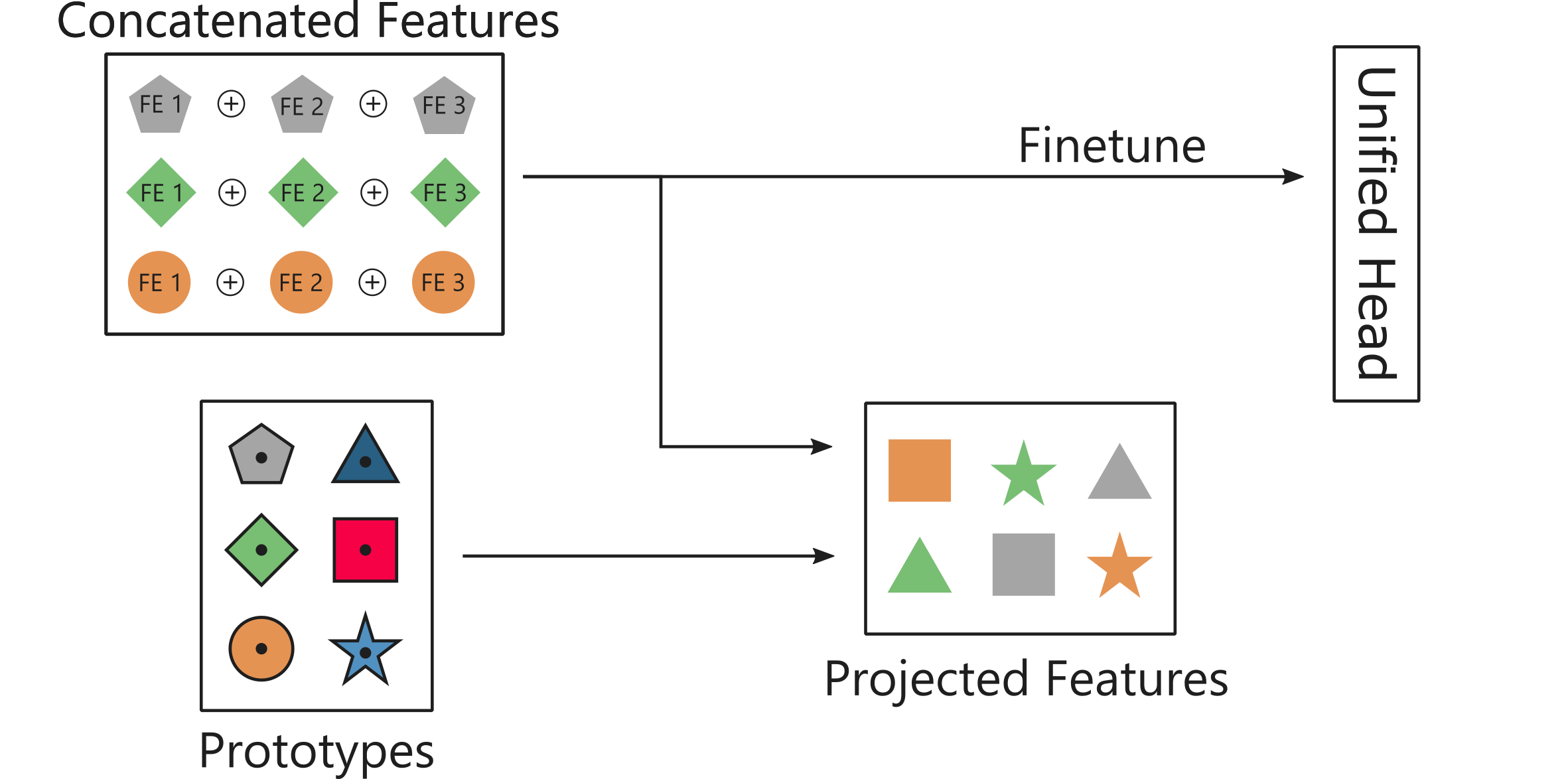}
        \caption{\textbf{Step 2}: Class prototypes are extracted or updated from the current task data. Incomplete class prototypes (those estimated before Step~1 extends or modifies a feature extractor) are updated and data for unavailable classes is simulated by pseudo-feature projection. An unbiased classification head is finetuned from the current training data and the projected features of unavailable classes.}
        \label{fig:step2_horde}
    \end{subfigure}
    \caption{Overview of the steps our proposed method (Horde) performs for each incremental task.}
    \label{fig:adjustment_overview}
\end{figure}

To effectively utilize this dynamic embedding space, we address the challenge of missing classes by constructing prototypes for all encountered classes. Class prototypes are used to train a unifying classification layer through an adjusted feature translation mechanism, termed \emph{pseudo-feature projection}, ensuring continuous adaptation and robust performance across all classes. Concretely, the learning approach is divided into two steps: \textbf{(1st)} based on the difficulty of the current task and depending on how well new classes can fit into the ensemble embedding space, the ensemble is expanded with a new feature extractor (described in \cref{sec:ensemble_fe}); \textbf{(2nd)} once the embedding space has been fixed for the current task, class prototypes are extracted and the unified classification layer is trained through the \emph{pseudo-feature projection} (described in \cref{sec:unified_head}). These steps are performed for every incremental task and are summarized in \Cref{fig:adjustment_overview}.

\subsection{Feature Extractor Ensemble}
\label{sec:ensemble_fe}
The proposed aggregation framework consists of multiple individual feature extractors (FEs), each trained on a specific task and then frozen to preserve the learned representations. The motivation for this zero-forgetting strategy is to enforce stability, avoiding any catastrophic forgetting on the ensemble while providing some plasticity through the extension of the ensemble. 
Unlike FeTrIL, which freezes a single feature extractor after the initial training, the extension of the feature space through the ensemble relieves the dependence of an expressive initial feature extractor. The goal of each feature extractor is to build a diverse and expressive feature space that emphasizes high-quality representations rather than optimizing the performance of the individual incremental task. Further, we adopt the self-learning loss from PASS~\cite{pass}. This self-learning loss enhances the learned feature representation by simultaneously classifying image orientation and categories (each image class now has 4 augmented labels depending on the image orientation). To further improve regularization on the feature space topology, we incorporate a metric learning head with contrastive loss~\cite{understanding_deep_learning} and hard-negative mining~\cite{hardnegativepair}. This promotes \emph{spherical-shaped} clusters in the embedding space, which improves class discrimination between known and unknown distributions~\cite{masana2018metric}. Additionally, the sphere-shaped structure aligns well with the properties of a multivariate Gaussian distribution, which relates to the pseudo-feature projection we propose. An ablation study of the effects of individual components is provided in the supplementary material (see \cref{sec:abl_fe_training}).

\minisection{Ensemble Growth.}
To control the growth of the ensemble, we set a predefined budget $B$ for the maximum number of FEs. For each incremental task, a decision is made whether the concatenated embedding space should be adjusted based on the following criteria:

\begin{itemize}
    \vspace{0.5em}
    \item \textit{constant feature representation}: when the current ensemble embedding representation is sufficient to handle the incremental task, no new FE is trained. New classes are learned using the existing ensemble representations without requiring additional feature extraction capacity.
    \vspace{0.5em}
    \item \textit{dynamic feature adaption}:
     when the current ensemble of FEs cannot adequately represent the new task due to a significant change in the data distribution, task complexity or overlap with previous classes, a new FE is added.
     \vspace{0.5em}
\end{itemize}

\noindent To capture these criteria and guide the growth of the ensemble, we propose two heuristics to guide the modification of the ensemble (see Step~1 in \cref{fig:step1_horde}):
\begin{itemize}
    \item \textbf{Class Set Maximisation ($\bm{\text{Horde}_m}$)}: this heuristic aims to maximize the diversity of classes represented across the ensemble. Specifically, it ensures that each FE contributes to representing as much of a distinct set of classes as possible
    \begin{equation}
\max\  \bigcup_{i\in B} \bigm|\! c\in F^{i}\!\bigm| \text{,}
\end{equation}
    thereby increasing the overall coverage of the class space across all feature extractors. This maximization is tested at the start of each incremental task. Thus, when a larger class set is possible with the current incremental task data, a new FE is trained. The new FE either is added or replaces one in the ensemble.
    \vspace{0.5em}
    
    \item \textbf{Task Error Rate ($\bm{\text{Horde}_c}$)}:
    At the start of the incremental task, the error rate~$e$ on the current incremental data is computed (before training). It is obtained from the confusion matrix~($\text{CM}$) by calculating the ratio of wrong predictions over all other predictions:

    \begin{equation}
      e \ =\  \frac{1}{|\mathcal{Y}|}\ \sum_{c\in\mathcal{Y}} \left(\ \frac{\sum_{j \neq c}^{} \text{CM}_{c, j}}{\sum_{i}^{} \text{CM}_{c, i}}\ \right)\text{.}
    \end{equation}

    If $e$ is too high, the incremental data cannot be classified with the current ensemble effectively.
    Therefore, we introduce a threshold or budget of the ensemble $B$ which signals the need to train a new feature extractor based on $e$. After training the unified head (Step~2), an \emph{improvement score} is calculated as the difference between the error rate at Step~1 (before any training is performed) and after Step~2. If the budget $B$ has been exceeded the FE with the lowest improvement score is replaced.
\end{itemize}

\subsection{Unified Classification Layer}
\label{sec:unified_head}
To unify the feature representations from the ensemble and enable task-agnostic classification, we utilize a fully-connected layer. This layer has dynamic input and output sizes depending on the growth of the ensemble and the number of incrementally learned classes. To mitigate task-recency bias~\cite{marc_survey2}, we train this unified head using both data from the current task and projected class prototype features through our proposed pseudo-feature projection.

\minisection{Pseudo-feature projection.} 
Pseudo-feature projection, inspired by FeTrIL~\cite{fetril}, extends feature translation by incorporating both the mean and standard deviation of class prototypes. This enhances the sampling of dimensions, reduces the chance of overlapping classes in the embedding space and leads to more accurate feature replay. With this projection, a data point from one class may be projected to a pseudo-feature representation of any other previously learned class. Our proposed projection extends the one from FeTrIL on \cref{eq:fetril-proj} as
\begin{equation}
\hat{F}_{c} = \mu_{c} + \frac{\mathbf{f}(\bm{x}_{i}; \theta) - \mu_{\bm{y}_{i}}}{\bm{\sigma}_{\bm{y}_{i}}} \cdot \bm{\sigma}_{c} \ \text{,}
\label{eq:horde-proj}
\end{equation}

\begin{table*}[ht!]
    \centering
    \adjustbox{max width=\textwidth}{
    \begin{tabular}{cllllllllllll}
        \toprule
        \textbf{Est. Method} & \textbf{T0} & \textbf{T1} & \textbf{T2} & \textbf{T3} & \textbf{T4} & \textbf{T5} & \textbf{T6} & \textbf{T7} & \textbf{T8} & \textbf{T9} & \textbf{T10} & \textbf{Avg. Acc. $\uparrow$} \\ \toprule
        zeros&   73.3 &   39.9 &   32.9 &   35.2 &   25.2 &   25.4 &   24.9 &   19.2 &   18.3 &   19.1 &   17.1 &   \ \ \ 30.0 \\ \midrule
        random-1.0&   73.3 &   39.9 &   32.9 &   35.2 &   25.2 &   25.4 &   24.8 &   19.2 &   18.3 &   19.0 &   17.2 &   \ \ \ 30.0 \\ 
        random-3.0&   73.3 &   47.0 &   37.9 &   38.0 &   29.0 &   31.3 &   30.0 &   22.9 &   21.9 &   22.5 &   22.5 &   \ \ \ 34.2 \\ 
        random-5.0&   73.3 &   52.5 &   42.7 &   44.4 &   33.6 &   34.5 &   33.8 &   25.2 &   25.9 &   26.3 &   26.4 &   \ \ \ 38.0 \\ 
        random-10.0&   73.3 &   59.2 &   50.8 &   52.7 &   43.5 &   43.4 &   40.2 &   31.4 &   31.8 &   31.8 &   31.7 &   \ \ \ 44.5 \\ 
        random-15.0&   73.3 &   61.7 &   53.5 &   54.2 &   45.6 &   46.1 &   42.0 &   35.8 &   \third{36.1} &   \third{34.8} &   34.4 &   \ \ \ 47.0 \\ 
        random-20.0&   73.3 &   62.5 &   55.5 &   \third{54.8} &   47.3 &   47.0 &   42.9 &   36.5 &   35.5 &   \third{34.8} &   34.4 &   \ \ \ 47.7 \\ 
        random-30.0&   73.3 &   \third{62.9} &   55.9 &   \second{55.1} &   \second{48.8} &   47.6 &   \third{43.7} &   36.8 &   36.0 &   \third{34.8} &   \third{34.7} &   \ \ \ \third{48.1} \\ 
        random-40.0 &   73.3 &   62.6 &   \second{58.0} &   \third{54.8} &   46.9 &   \third{47.8} &   \second{44.9} &   \second{38.6} &   \second{36.5} &   \second{35.9} &   \second{34.9} &   \ \ \ \second{48.6} \\ 
        random-50.0&   73.3 &   \second{64.1} &   57.2 &   \third{54.8} &   \third{47.6} &   \second{47.9} &   42.0 &   \third{37.6} &   33.8 &   \third{34.8} &   34.4 &   \ \ \ 48.0 \\ 
        random-75.0&   73.3 &   62.1 &   \third{57.9} &   54.5 &   46.2 &   45.4 &   40.7 &   35.6 &   34.1 &   33.7 &   33.6 &   \ \ \ 47.0 \\     
        random-100.0&   73.3 &   62.6 &   56.5 &   53.9 &   46.5 &   45.8 &   40.7 &   34.4 &   34.2 &   33.8 &   33.4 &   \ \ \ 46.8 \\ \midrule
        original features &   73.3 &  \best{64.3} &  \best{61.3} &  \best{60.8} &  \best{55.1} &  \best{53.7} &  \best{52.7} &   \best{46.8} &   \best{47.1} &   \best{46.3} &   \best{45.2} &   \ \ \ \best{55.2} \\ \bottomrule
    \end{tabular}}
    \caption{Results for class prototype estimation when the corresponding class prototype is not available during training. The evaluation is performed on a CIL 50/10 setup with Slim-Resnet-18 only. \best{1st}, \second{2nd} and \third{3rd} best metrics are marked accordingly.}
    \label{tab:results_ablation_estimate}
\end{table*}

\noindent where $\hat{F}_{c}$ represents the pseudo-features of the latent representation of a data point $(\bm{x}_{i},\bm{y}_{i})$ which is projected from the original class $\bm{y}_{i}$ to the desired class $c$. This transformation leverages the class prototypes; specifically the mean~$\mu_{\bm{y}_{i}}$ and standard deviation~$\sigma_{\bm{y}_{i}}$ to modify the latent representation $\mathbf{f}(\bm{x}_{i}; \theta)$. Class prototypes are updated during Step~2, before training the unified classification layer and after the ensemble has been adjusted.

We represent a complete class prototype as the concatenation of the individual class statistics from each FE in the ensemble:
\begin{equation}
\begin{split}
    \bm{\mu}_c &= (\bm{\mu}_{c, 1}, \, \dots \,, \bm{\mu}_{c, n})\text{,}\\
    \bm{\sigma}_c &= (\bm{\sigma}_{c, 1}, \, \dots \,, \bm{\sigma}_{c, n})\text{,}
\end{split}
\end{equation}

\noindent where $n$ determines the current size of the ensemble. Throughout the incremental sequence, the ensemble can be expanded until the feature extractor budget is exhausted ($n\!\leq\!B$). Once this limit has been reached, individual feature extractors need to be finetuned or replaced and their corresponding class prototype ($\bm{\mu}_{c, i}$, $\bm{\sigma}_{c, i}$) is reset.

Class prototypes of certain classes may be incomplete for newly added or modified FEs. When class statistics are unknown for a specific FE, estimates are required for pseudo-feature projection to calculate $\bm{\hat{\mu}}_{c, \mathbf{f}}$ and $\bm{\hat{\sigma}}_{c, \mathbf{f}}$. In the absence of statistical information, we fix the standard deviation to $\bm{\hat{\sigma}_{c, \mathbf{f}}}\!=\!\bm{1}$. This decision is based on the fact that the estimation of $\bm{\hat{\mu}}_{c, \mathbf{f}}$ already provides sufficient variance. Therefore, for the estimation of the mean component $\bm{\hat{\mu}}_{c, \mathbf{f}}$ we propose three heuristics:
\vspace{0.5em}
\begin{enumerate}
    \item \textbf{zeros}: clamping all $\bm{\hat{\mu}}_{c, \mathbf{f}}$ estimations to $0$
    \begin{equation}
     \bm{\hat{\mu}}_{c, \mathbf{f}} = \bm{0}\,\text{,}
    \end{equation}
    \item \textbf{random}: randomly sample $\bm{\hat{\mu}}_{c, \mathbf{f}}$ from a multivariate normal distribution
    \begin{equation}
     \bm{\hat{\mu}}_{c, \mathbf{f}} \sim \mathcal{N}(\bm{0}; \bm{\Sigma})\,\text{,}
    \end{equation}
    \item \textbf{original features}: estimate $\bm{\hat{\mu}}_{c, \mathbf{f}}$ with the original representation of the transforming sample and use them without modification 
    \begin{equation}
     \bm{\hat{\mu}}_{c, \mathbf{f}} = \mathbf{f}(\bm{x}_{i}; \theta)\,\text{.}
    \end{equation}
\end{enumerate}

We evaluate the proposed feature estimation heuristics in an empirical experiment on a class-incremental learning scenario with no repetition. The results on CIFAR 50/10 trained on a Slim-Resnet-18 are listed in \Cref{tab:results_ablation_estimate} (see \cref{sec:experiments} for more details). This scenario requires the estimation of class prototype components (e.g., mean, variance) at each incremental task and the estimation is essential for the classification. The \emph{original features} estimation performed best, and this heuristic is the one used in all subsequent experiments.

In EFCIR scenarios, the repetition of classes within incremental tasks enhances the performance of pseudo-feature projection as it aligns individual FE representation spaces by eliminating the need for estimating class prototype components. During class repetition they can be directly calculated from the available task.

\section{Experimental Setup}
\label{sec:experiments}
Most incremental learning methods expect a different set of classes with all dataset samples for each class available when learning its corresponding task.
However, when class repetition is introduced, the complexity of potential scenarios increases significantly, and where sequence length and repetition frequency become additional variables. To address this, we propose an analysis into the effects of class repetition within a setting that shares many characteristics of traditional incremental learning but incorporates longer sequences with class repetition.
Code for the proposed scenarios and methods is available\footnote{ \mbox{\url{www.github.com/Tsebeb/cvww\_cir\_horde}}}.

Overall, the proposed experiments aim to analyze a) the performance of IL methods in scenarios without repetition (baseline), b) the performance of CIL with small incremental tasks and class repetition, and c) the resilience of the methods against bias deviations in repetition frequency.

Ideally, we expect the average accuracy of our proposed method to be on par with state-of-the-art methods on (a) and to outperform them in (b) and (c). To validate this, method performance will be ranked based on average accuracy for all scenarios (a -- c).

\subsection{Compared Methods}
We benchmark a total of 14 methods, which include two rehearsal-based approaches, five incremental learning methods, five state-of-the-art exemplar-free class-incremental learning (EFCIL) methods, and two variants of our proposed approach. The two rehearsal-based methods are excluded from the ranking and serve as an upper baseline (Joint~\cite{marc_survey2}) and a reference point (Weight-Alignment~(WA)~\cite{weigth_align}; $n=2000$).

The five incremental learning methods consist of two baseline methods (Freezing~(FZ) and Finetuning (FT)~\cite{marc_survey1}), and three classic IL methods Elastic Weight Consolidation~(EWC)~\cite{ewc}, Memory Aware Synapses~(MAS)~\cite{mas} and Learning without Forgetting~(LWF)~\cite{lwf}. These three methods were not originally proposed for CIL, thus, requiring the use of a task-ID at inference time. However, they are easily and commonly adaptable to task-agnostic settings. As such, we performed a grid search for their optimal hyperparameters based on the CIL 50/10 setting and used these for the repetition settings.

The five state-of-the-art, rehearsal-free, protoype-based methods comprise:  Prototype Augmentation and Self-Supervision~(PASS)~\cite{pass}, Class-Incremental Learning with Dual Supervision~(IL2A)~\cite{il2a}, Self-Sustaining Representation Expansion~(SSRE)~\cite{ssre}, Prototype Reminiscence and Augmented Asymmetric Knowledge Aggregation~(PRAKA)~\cite{praka} and Feature Translation for Exemplar-free Class Incremental Learning~(FeTrIL)~\cite{fetril}. These methods were originally reported on the CIL CIFAR 50/10 setting. Therefore, since the proposed repetition scenarios are closely related to this setting, we use the hyperparameters proposed by the original authors.

Finally, we evaluate our proposed method with both ensemble growth heuristics ($\text{Horde}_\text{m}$ and $\text{Horde}_\text{c}$).
A detailed overview of the used hyperparameters is provided in the supplementary material (see \cref{sec:hyperparameters}).

\subsection{Model Architecture}
All methods employ the same base feature extractor, a ResNet-18~\cite{resnet} model that has been adjusted to the CIFAR input dimensions~\cite{pass, il2a}.
For our approach, which utilizes an ensemble of feature extractors, we employ a slimmed-down variant of ResNet-18 for incremental tasks. This variant reduces the number of channels/filters for convolutions while preserving the network’s depth (see supplementary material \cref{sec:architecture}). With this reduced architecture, we construct an ensemble consisting of one full ResNet-18 and nine Slim-ResNet-18 models (making our budget $B\!=\!10$). This configuration results in a total number of parameters and computational requirements (see \Cref{tab:model_size}) that are roughly equivalent to those of knowledge distillation approaches (or importance weight estimation~\cite{ewc, mas, liu2018rotate}).

\subsection{Scenarios}
Experiments are conducted on the CIFAR-100 dataset~\cite{cifar}, employing data augmentation in line with other CIL methods~\cite{il2a, praka}. These augmentations consist of a 4-pixel zero padding of the input image and a random cropping to the original $32\times32$ size. Followed by a random horizontal flip, image brightness jitter and image normalization.

To evaluate the effects of repetition on CIL methods, we organize the experiments in three scenarios. First, a baseline is established by evaluating (a) all methods on an incremental learning scenario without repetition.

\vspace{.3em}
\begin{enumerate}[label=(\alph*)]
    \item \textbf{CIL 50/10.} The classic task-agnostic class-incremental scenario consisting of an initial training session with 50 classes and followed by 10 incremental tasks, each containing five novel classes.
\end{enumerate}
\vspace{.3em}

\noindent Second, we evaluate (b) performance on a modified CIL scenario where classes repeat in the task sequence. Specifically, the scenario is built by replacing the discrete incremental tasks with clear boundaries from CIL 50/10 with small (2,000 training samples per task) incremental tasks that can contain class repetition. Each class, old or new, has the same probability of being in an incremental task.

\vspace{.3em}
\begin{enumerate}[label=(\alph*)]
\setcounter{enumi}{1}
    \item \textbf{EFCIR-U 50/100.} Similarly to the CIL 50/10 scenario, the initial training also covers 50 classes. An essential element of repetition is a mixture of new and already seen samples. Therefore, we only provide 50\% of the available training data samples for the initial training. Following the initial task, the scenario consists of 99 small, incremental tasks, with a limit of 2,000 training samples each. Both the initial 50 and incremental 50 classes have a fixed probability of $15\%$ of being discovered or repeated in an incremental task so that tasks do not contain too many classes on average. The number of samples per class in a task are balanced as in the CIL 50/10 scenario.
\end{enumerate}
\vspace{.3em}

\noindent In the third scenario, the aim is to assess the IL method's (c) resilience against biases in repetition frequency. To establish this bias during scenario creation we propose to draw the repetition probability of each class from a \emph{Beta Distribution}~\cite{beta_distribution}. An illustration of the repetition bias is provided in the supplementary material \cref{sec:scenario_viz}. 

\vspace{.3em}
\begin{enumerate}[label=(\alph*)]
\setcounter{enumi}{2}
    \item \textbf{EFCIR-B 50/100.} To test the resiliency against repetition frequency, we sample individual class repetition probabilities $p\!\sim\!\text{Beta}(\alpha, \beta)$ with parameters $\alpha\!\!=\!\!3.5$ and $\beta\!\!=\!\!20.0$. This way, the expectation \mbox{$\mathbb{E}[\text{Beta}(3.5, 20.0)]\!\approx\!0.15$} is similar to the uniform EFCIR-U scenario, implying that on average the same number of classes are present in each task.
\end{enumerate}
\vspace{.3em}

\begin{table}[t]
    \centering
    \begin{tabular}{cc}
        \toprule
         \textbf{Model} & \textbf{\# Parameters} \\ \midrule
         ResNet-18 & 11.307.956 \\ 
         Slim ResNet-18 &  1.109.240\\ \midrule
         Knowledge Distillation & 22.615.912 \\
         Ensemble (ours) & 21.291.116 \\ \bottomrule
    \end{tabular}
    \caption{Number of parameters for different architectures.}
    \label{tab:model_size}
\end{table}

\noindent In scenarios with repeated classes, the optimal learning rate and number of epochs depend on various factors (\emph{e.g}\onedot method, number of training samples, length of incremental sequence) and are highly influential. To address this, we split 10\% of the available training data as a validation set. For all methods, we apply early stopping~\cite{astonpr524, understanding_deep_learning} using this validation data for the classes present in the task. We monitor the validation loss (including regularization and auxiliary losses of the method) and allow for a patience period of 5 epochs. If no improvement is observed, we perform a learning rate decay step. Each decay step reduces the learning rate by a factor of $0.1$, the model weights are reset to the best checkpoint before patience, and we do not perform more than 2 decay steps.

\minisection{Evaluation.} All scenarios are ranked by the average accuracy~\cite{marc_survey1, marc_survey2,icarl,fetril,pass} achieved over the complete task sequence. Average accuracy is calculated by evaluating the model on the CIFAR test set based on the classes that have been seen up to each task. Complementary to the average accuracy, average \emph{forgetting}~\cite{rwalk, marc_survey1, marc_survey2} is also reported, which measures the drop in accuracy over the task sequence. Experimental results are averaged over 5 seeds. 

\begin{table*}[ht!]
    \centering
    \resizebox{0.9\linewidth}{!}{
    \begin{tabular}{c@{\hskip 0.3in}cc@{\hskip 0.3in}cc@{\hskip 0.3in}cc}
         & \multicolumn{2}{@{\hskip -0.3in}c}{\textbf{(a) CIL 50/10}} & \multicolumn{2}{@{\hskip -0.3in}c}{\textbf{(b) EFCIR-U 50/100}} & \multicolumn{2}{c}{\textbf{(c) EFCIR-B 50/100}} \\ \toprule
        \textbf{Method} & \textbf{Avg. $\bm{A \uparrow}$} & \textbf{Avg. $\bm{f \downarrow}$} & \textbf{Avg. $\bm{A \uparrow}$} & \textbf{Avg. $\bm{f \downarrow}$} & \textbf{Avg. $\bm{A \uparrow}$} & \textbf{Avg. $\bm{f \downarrow}$} \\ \toprule
        Joint &  73.9 & - & 69.8 & - & 68.9 & - \\ 
        WA~\cite{weigth_align} & \spacerbullet{42.7 $\pm$ 2.3} & \spacerbullet{33.2 $\pm$ 1.4} & \spacerbullet{50.4 $\pm$ 0.2} & \spacerbullet{16.7 $\pm$ 2.4} & \spacerbullet{49.2 $\pm$ 0.7} & \spacerbullet{18.0 $\pm$ 1.6} \\ \midrule
        FT & \spacerbullet{14.2 $\pm$ 1.0} & \spacerbullet{ 57.8 $\pm$ 1.2} & \spacerbullet{36.2 $\pm$ 2.1} & \spacerbullet{25.6 $\pm$ 2.7} & \spacerbullet{34.2 $\pm$ 2.0} & \spacerbullet{29.0 $\pm$ 2.6} \\
        FZ & \spacerbullet{52.6 $\pm$ 1.4} & \spacerbullet{19.7 $\pm$ 0.9} & \spacerbullet{40.2 $\pm$ 3.9} & \spacerbullet{20.0 $\pm$ 1.6} & \spacerbullet{41.7 $\pm$ 3.1} & \spacerbullet{22.5 $\pm$ 1.8} \\
        EWC~\cite{ewc} & \spacerbullet{45.9 $\pm$ 2.9} & \spacerbullet{25.7 $\pm$ 1.4} & \spacerbullet{47.7 $\pm$ 3.2} & \second{13.5 $\pm$ 1.5} & \spacerbullet{45.5 $\pm$ 3.2} & \third{17.8 $\pm$ 1.8} \\
        MAS~\cite{mas} & \spacerbullet{45.9 $\pm$ 2.9} & \spacerbullet{25.8 $\pm$ 1.4} & \third{49.3 $\pm$ 2.6} & \best{12.0 $\pm$ 1.8} & \third{47.2 $\pm$ 2.3} & \best{16.1 $\pm$ 2.1} \\
        LwF~\cite{lwf} & \spacerbullet{47.9 $\pm$ 1.8} & \spacerbullet{24.1 $\pm$ 0.8} & \spacerbullet{45.7 $\pm$ 1.9} & \third{15.9 $\pm$ 2.8} & \spacerbullet{43.5 $\pm$ 0.8} & \spacerbullet{19.8 $\pm$ 4.1} \\ \midrule
        PASS~\cite{pass} & \spacerbullet{62.1 $\pm$ 1.9} & \spacerbullet{14.1 $\pm$ 0.4} & \spacerbullet{30.2 $\pm$ 2.0} & \spacerbullet{35.3 $\pm$ 2.1} & \spacerbullet{30.6 $\pm$ 1.4} & \spacerbullet{38.3 $\pm$ 1.6} \\
        PRAKA~\cite{praka} & \best{63.1 $\pm$ 2.5} & \best{11.8 $\pm$ 2.2} & \spacerbullet{43.1 $\pm$ 2.1} & \spacerbullet{22.3 $\pm$ 2.3} & \spacerbullet{42.7 $\pm$ 3.2} & \spacerbullet{25.6 $\pm$ 1.8} \\
        IL2A~\cite{il2a} & \spacerbullet{54.2 $\pm$ 1.4} & \spacerbullet{19.1 $\pm$ 1.3} & \spacerbullet{26.3 $\pm$ 3.0} & \spacerbullet{32.2 $\pm$ 2.9} & \spacerbullet{27.2 $\pm$ 2.5} & \spacerbullet{37.2 $\pm$ 1.7} \\
        SSRE~\cite{ssre} & \spacerbullet{53.0 $\pm$ 2.7} & \second{13.0 $\pm$ 0.8} & \spacerbullet{29.2 $\pm$ 3.5} & \spacerbullet{25.4 $\pm$ 2.1} & \spacerbullet{26.5 $\pm$ 2.2} & \spacerbullet{26.4 $\pm$ 2.1} \\
        FeTrIL~\cite{fetril} & \spacerbullet{61.4 $\pm$ 0.4} & \third{13.6 $\pm$ 0.8} &  \spacerbullet{46.5 $\pm$ 0.7} & \spacerbullet{22.9 $\pm$ 0.7} & \spacerbullet{46.9 $\pm$ 0.9} & \spacerbullet{23.8 $\pm$ 1.2} \\ \midrule
        $Horde_m$ & \second{62.9 $\pm$ 1.2} & \spacerbullet{15.2 $\pm$ 0.7} & \best{54.4 $\pm$ 0.7} & \spacerbullet{16.4 $\pm$ 1.5} & \best{54.3 $\pm$ 0.4} & \second{17.7 $\pm$ 1.0} \\
        $Horde_c$ & \third{62.9 $\pm$ 1.2} & \spacerbullet{15.3 $\pm$ 0.6} & \second{53.4 $\pm$ 0.7} & \spacerbullet{17.6 $\pm$ 1.6} & \second{53.1 $\pm$ 0.4} & \spacerbullet{18.5 $\pm$ 1.1} \\ \bottomrule
    \end{tabular}
    }
    \caption{Average Accuracy (Avg. $A$) and average Forgetting (Avg. $f$) for all 3 proposed scenarios. The listed results are averaged over 5 seeds (except incremental Joint). The 3 best results are marked with a \best{gold}, \second{silver} and \third{bronze} medal respectively.}
    \label{tab:res_aggregate}
\end{table*}

\section{Results}
The summarized results of the experiments are listed in \Cref{tab:res_aggregate}. Detailed plots and tables of the accuracy progression for all methods in each proposed scenario are provided in the supplementary material (\cref{sec:scenario_results}).

\minisection{Scenario (a) CIL 50/10.}
The conducted baseline experiment confirms the reported results from other works~\cite{marc_survey2, pass, il2a, praka}. We observe a significant performance gain of approximately $10\!\mbox{ - }\!15\%$ in average accuracy over the task sequence when comparing the state-of-the-art rehearsal-free (EFCIL) methods with EWC, MAS and LwF. While our proposed method is particularly designed towards repetition scenarios, where the estimation of class prototype components is not always required, it remains competitive in disjoint, no-repetition scenarios as well, showing comparable performance to the best EFCIL methods~\cite{praka, pass, il2a, fetril}.

\begin{figure}[t]
    \centering
    \includegraphics[width=.95\linewidth, trim=18 10 5 18, clip]{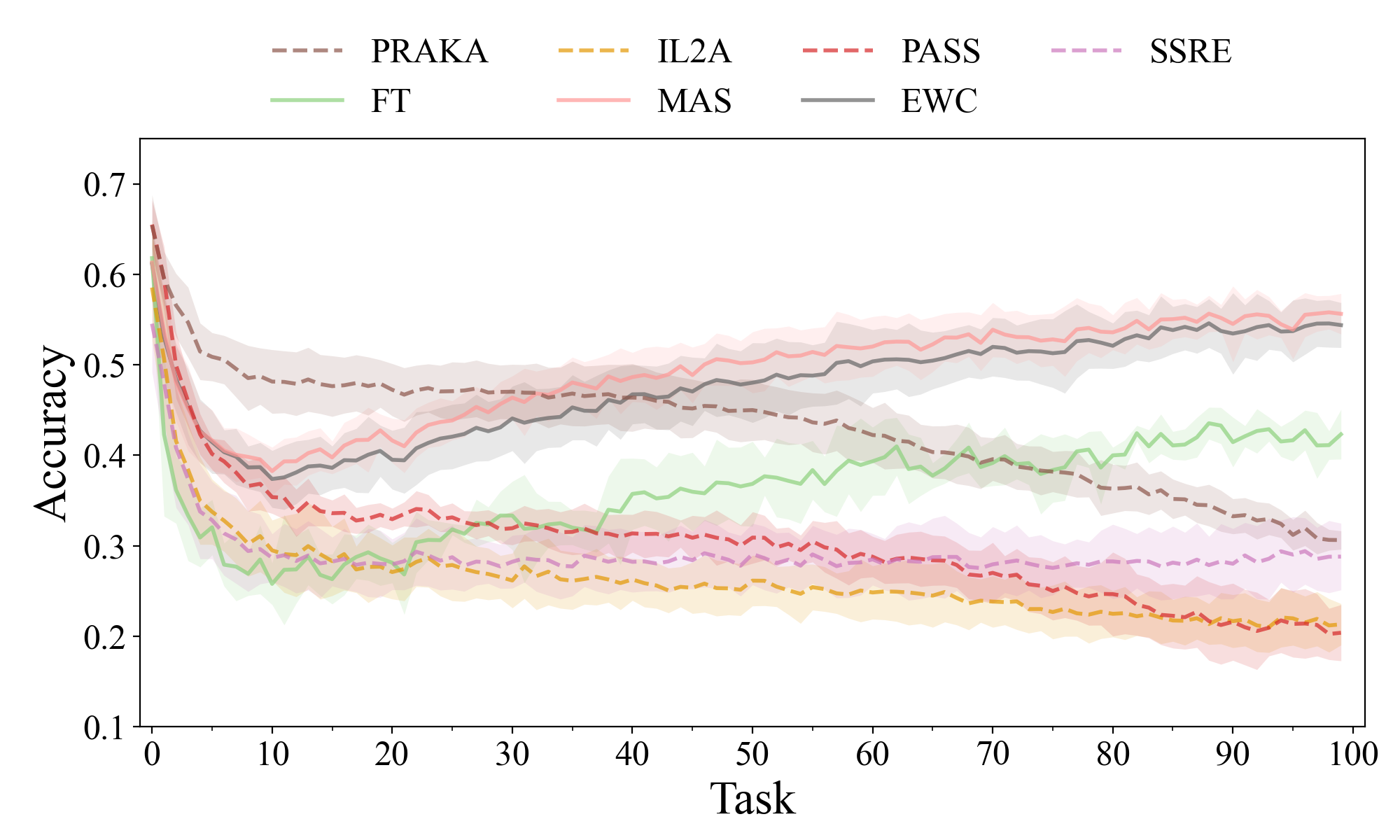}
    \caption{Accuracy curves for scenario (b) with equiprobable repetition frequency. Weight-regularized methods (\textit{solid}) benefit directly from short tasks with class repetition, while prototype-based approaches (\textit{dashed}) degrade in accuracy as the sequence advances.}
    \label{fig:efcir_uniform_weight_kd}
\end{figure}

\minisection{Scenario (b) EFCIR-U.}
Introducing class repetition in small incremental tasks into the scenario leads to significant performance differences. Weight-regularization approaches and vanilla finetuning typically underperform compared to knowledge distillation or class prototype-based approaches in EFCIL~\cite{praka}. However, in this scenario with repetition, we observe greatly improved performance for FT, EWC and MAS. The results for these methods surpass even the results from the CIL 50/10 scenario by leveraging data repetition effectively (see \cref{fig:efcir_uniform_weight_kd}). In contrast, EFCIL methods (PASS, IL2A, SSRE, PRAKA) that rely on both class prototype rehearsal and knowledge distillation show a performance degradation under repetition. This decline is not observed in methods that use either knowledge distillation (LwF) or class prototype rehearsal with frozen feature extractors (FeTrIL, Ours). 

\begin{figure*}[t]
    \centering
    \begin{subfigure}[c]{0.45\linewidth}
        \includegraphics[width=1.0\linewidth]{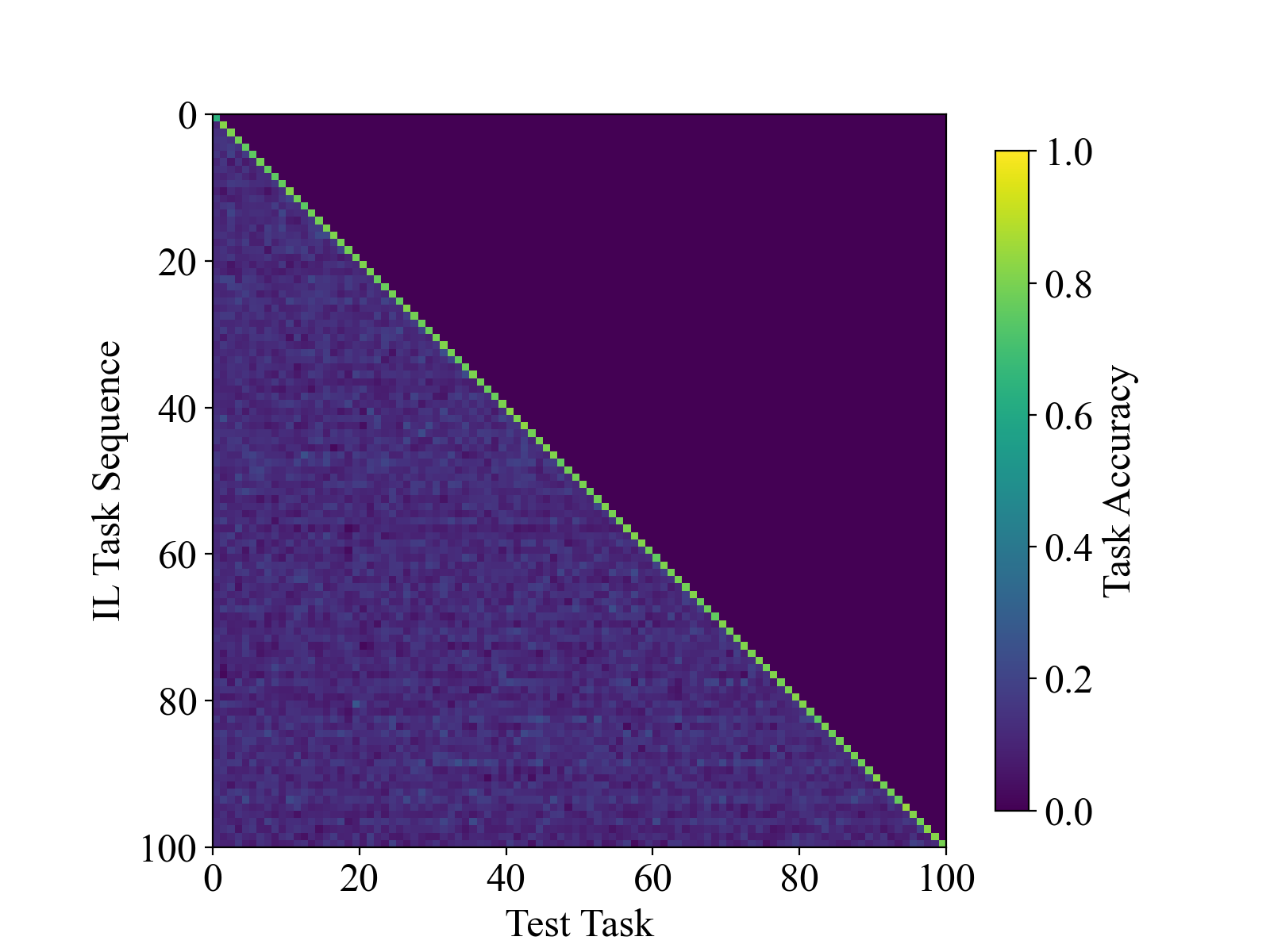}
        \caption{Cross-entropy loss gradient applied to all class weights in the classification head. Significant task recency bias is visible (the diagonal is significantly higher, with sharp drops after each task).}
    \end{subfigure}
    \hfill
    \begin{subfigure}[c]{0.45\linewidth}
        \includegraphics[width=1.0\linewidth]{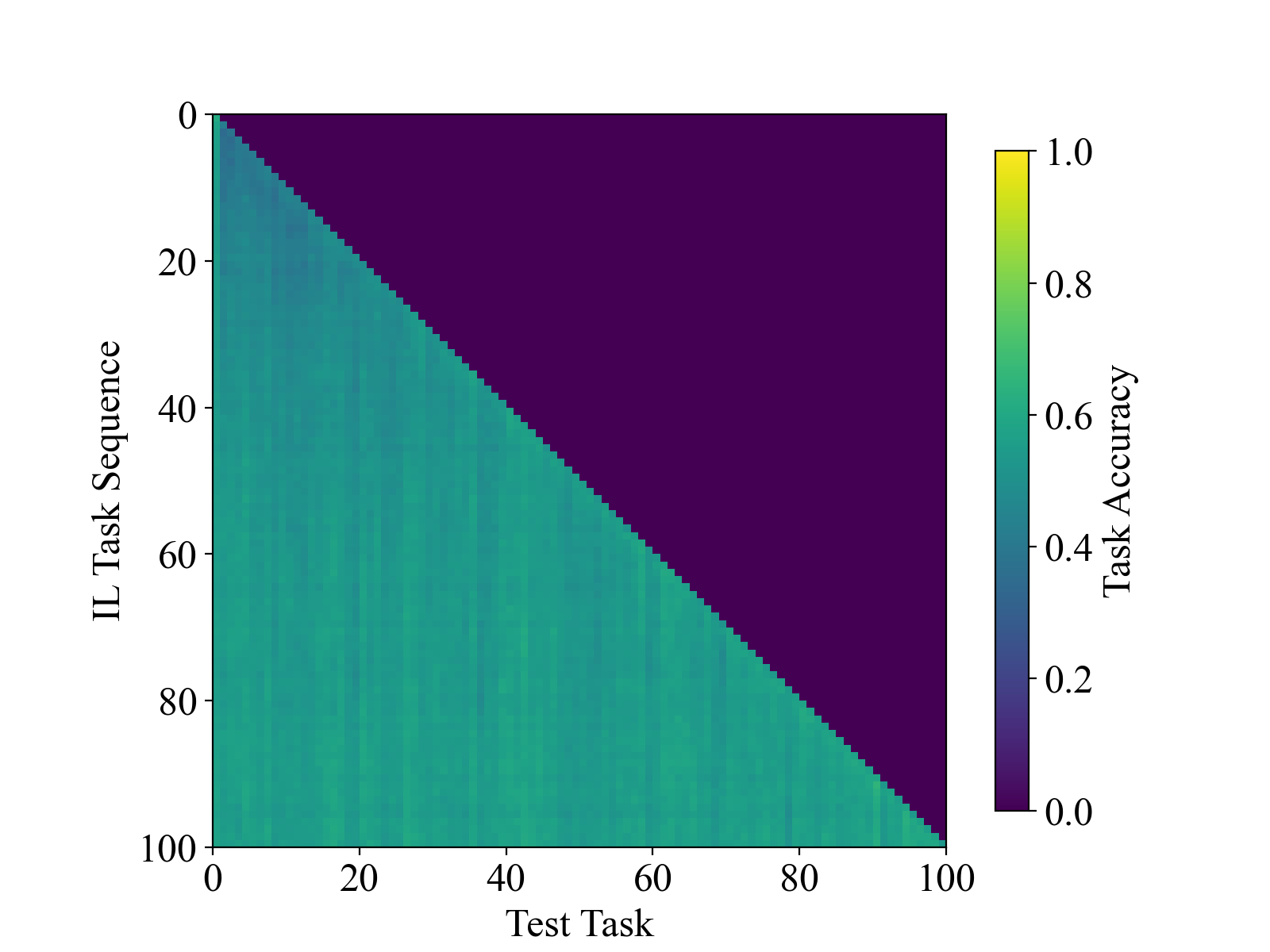}
        \caption{Cross-entropy loss gradient applied to only current task class weights in the classification head. Much of the task recency bias is alleviated by freezing the classifier weights for unavailable classes.}
    \end{subfigure}
    \caption{Depiction of the task accuracy progression of MAS over the scenario (b) sequence (averaged over 5 seeds). Accuracy is evaluated on the test set for the classes represented in the corresponding incremental training data within a task. Note, that for repetition there is always a certain overlap within tasks.}
    \label{fig:acc_triangle_task}
\end{figure*}

We hypothesize that the estimation of class prototypes with incomplete class data distribution in the former methods leads to a suboptimal feature embedding space, which is then propagated through the incremental task sequence via knowledge distillation. Frozen feature extractors, on the other hand, avoid this issue since their representations remain fixed after the initial training, preventing catastrophic drift in the embedding space during the task sequence.
This raises the question whether the assumption that the complete training data distribution of an individual class -- as in traditional class-incremental learning -- is a realistic assumption for continual learning scenarios.

Our ensemble-based approach (Horde), with both ensemble growth heuristics, establishes a new state-of-the-art for the repetition settings. Notably, when comparing our method with the closely related FeTrIL approach, we observe a performance increase under repetition. This suggests that our approach could extend the feature space of the base feature extractor by incorporating class combinations from smaller feature extractors. Over time, repetition aligns these representations, enabling the model to learn a unified classification head on a more diverse representation space provided by the ensemble.

\begin{figure}
    \centering
    \includegraphics[width=\linewidth, trim=18 13 7 18, clip]{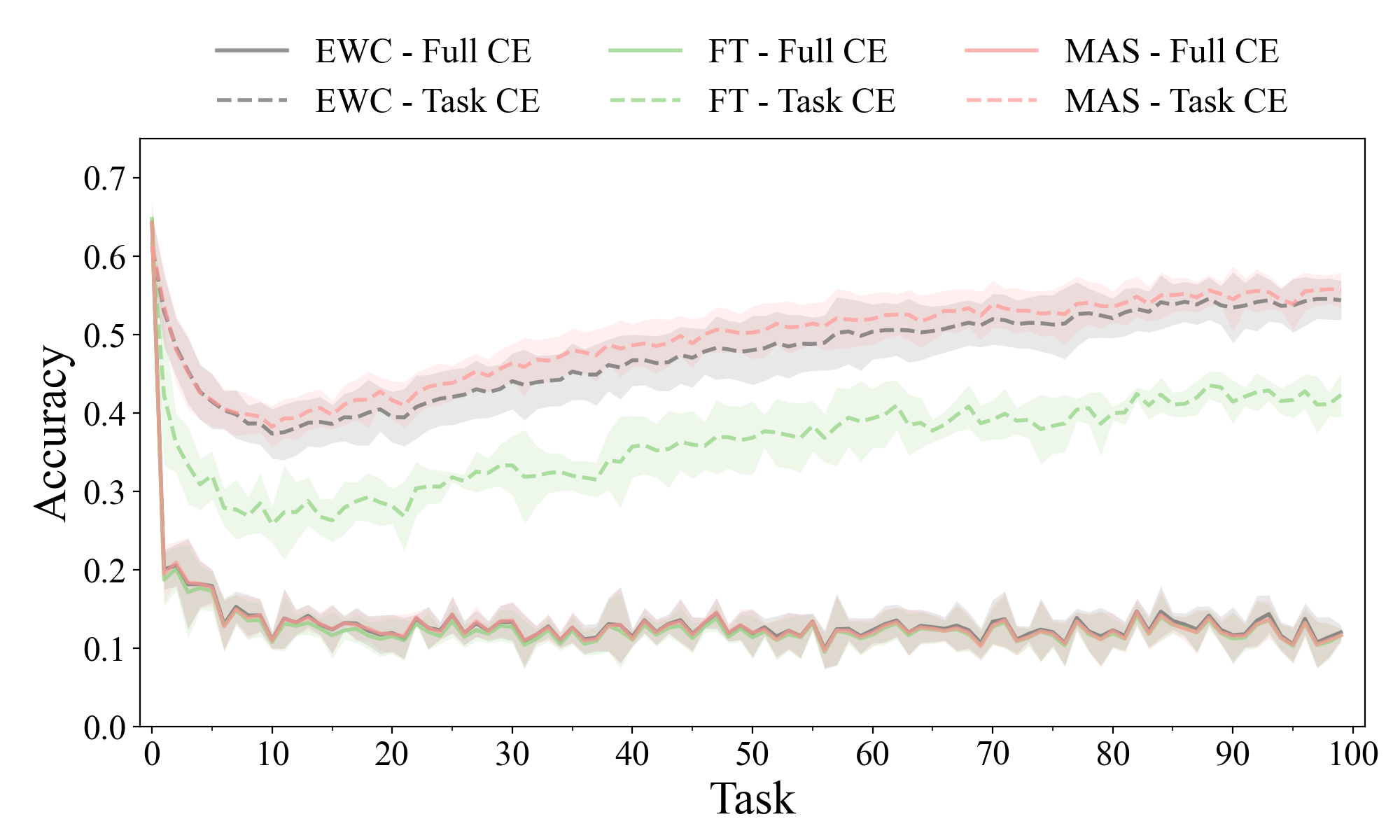}
    \caption{Accuracy results for finetuning and weight-regularization based methods. Solid lines indicate the backpropagation of the cross-entropy loss over all classes leading to catastrophic class recency bias. Dashed lines indicate the freezing of the weights related to the output of non-current classes.}
    \label{fig:abl_task}
\end{figure}

The strong performance gains for weight-regularized approaches are only observed when the cross-entropy loss during training is limited to the classes that are present within the current task. Practically, this is achieved by freezing all weights associated with classes outside of the current task~\cite{marc_survey1}. \Cref{fig:abl_task,fig:acc_triangle_task} illustrate the consequences of backpropagating the loss through all weights of the classification head. In this case, regardless of whether weight-regularization is applied to the feature extractor, a strong class recency bias emerges in the classification head. As a result, the accuracy of all three methods collapses, with the model essentially forgetting classes proportionally to how long they were seen last (see \cref{fig:acc_triangle_task}).
However, when the weights for unavailable classes in the classification head are frozen, a significant performance improvement is observed, as the model retains its ability to distinguish across earlier tasks without being overly biased towards the most recent ones.

\minisection{Scenario (c) EFCIR-B.}
The bias in repetition frequency appears to have only a minor effect on the average accuracy of the approaches. All tested methods achieve similar results or experience only a slight drop of up to $\sim\!\!2\%$ in average accuracy. This suggests that repetition frequency bias is a relatively minor challenge in the EFCIR-B 50/100 scenario. However, it is important to note that this setting only evaluates adjustments in repetition frequency while the sample distribution within a training task is kept balanced. Therefore, further investigation is needed to assess whether an imbalanced training data distribution in conjunction with biased repetition frequency would increase the difficulty. We leave this exploration to future work.

\section{Conclusion}
In this work, we conducted an exploratory evaluation of CIL methods in exemplar-free class-incremental learning with repetition scenarios and investigated their resiliency to biases in the repetition frequency of classes.

In the evaluated repetition scenarios, EFCIL methods that rely on class prototypes (PASS, PRAKA, IL2A, SSRE) severely underperform and are unable to benefit from the repetition of classes. Notably, weight-regularization-based approaches perform exceptionally well in repetition scenarios provided that training with cross-entropy is restricted to the classes present in each task, thereby mitigating the risk of class-recency bias in the classification head. 
The results from the repetition frequency bias from a beta distribution show only minimal performance differences, with either no effect on average accuracy or a slight drop of up to 2\%. 
Thus, a bias in repetition frequency alone without a biased sample distribution within a training task is insufficient for significant classification bias.

Furthermore, we introduce a novel ensemble learning technique that takes advantage of class repetition. This method combines a dynamic set of independent feature extractors, which are aligned through a unified head in a process we call pseudo-feature projection. The proposed method demonstrates competitive performance in traditional no-repetition settings and establishes a new state-of-the-art for scenarios with repetition.

\section*{Acknowledgements}
Marc Masana acknowledges the support by the “University SAL Labs” initiative of Silicon Austria Labs (SAL).

{
    \small
    \bibliographystyle{ieeenat_fullname}
    \bibliography{references}
}

\maketitlesupplementary
\setcounter{section}{0}
\def\thesection{\Alph{section}}

\section{Pseudo Code}
The algorithm for the proposed Horde method can be separated into two parts: (1st) the training of an individual feature extractor (FE), which is listed in \cref{alg:horde-fe-training} and (2nd) the overall assembly of the ensemble and training of the unified head through pseudo-feature projection (see \cref{alg:horde-unified}). The training of a feature extractor (1st part) can be freely adjusted (loss, network architecture) as long as a frozen feature extractor that can produce an embedding is the result.

\begin{algorithm}
    \centering
    \caption{FE Training}
    \label{alg:horde-fe-training}
    \begin{algorithmic}[1] 
        \State Initialize CE and ML Head
        \State Initialize new FE (or transfer learned weights)
        \For{training epoch}
            \For{$\bm{X}; \bm{Y}$ in Dataloader}
                \State $\bm{X}; \bm{Y} \leftarrow \text{SelfSupervision}(\bm{X}; \bm{Y})$
                \State Extract $\bm{\hat{X}} \leftarrow \text{FE}(\bm{X})$
                \State Predict $\bm{\hat{Y}} \leftarrow \text{Head}_{\text{CE}}(\bm{\hat{X}})$
                \State Project $\bm{A} \leftarrow \text{Head}_{\text{ML}}(\bm{\hat{X}})$
                \State Calculate $\mathcal{L}_{\text{CE}}$ (from $\bm{Y}$ and $\bm{\hat{Y}})$
                \State Calculate $\mathcal{L}_{\text{ML}}$ (with Hard Neg. Pairs on $\bm{A}$)
                \State Backprop $\mathcal{L}_{\text{CE}} + \mathcal{L}_{\text{ML}}$
            \EndFor
        \EndFor
        \State Remove CE and ML head
        \State Freeze FE
    \end{algorithmic}
\end{algorithm}

\begin{algorithm}
    \centering
    \caption{IL through pseudo-feature projection}
    \label{alg:horde-unified}
    \begin{algorithmic}[1] 
        \For{task}
            \If{Growth Condition ($\bm{\text{Horde}_m}$ or $\bm{\text{Horde}_c}$)} 
                \State Train FE (Algorithm~\ref{alg:horde-fe-training})
                \State Add / Replace FE in ensemble
            \EndIf
            \State Calculate $\bm{\mu}_c$ and $\bm{\sigma}_c$ for all current classes $c$
            \For{training epoch} \Comment{Only Unified head}
                \For{Batch}
                    \State Calculate $\mathcal{L}_{\text{CE}}$
                    \State Generate $\hat{F}_{c}$ from current Batch
                    \State Calculate $\mathcal{L}_{\text{CE};\text{P}}$ for $\hat{F}_{c}$
                    \State Backprop $\mathcal{L}_{\text{CE}} + \mathcal{L}_{\text{CE};\text{P}}$
                \EndFor
            \EndFor
        \EndFor
    \end{algorithmic}
\end{algorithm}

\section{Details about the Model Architecture}
\label{sec:architecture}
\begin{figure}[t]
    \centering
    \includegraphics[width=.85\linewidth]{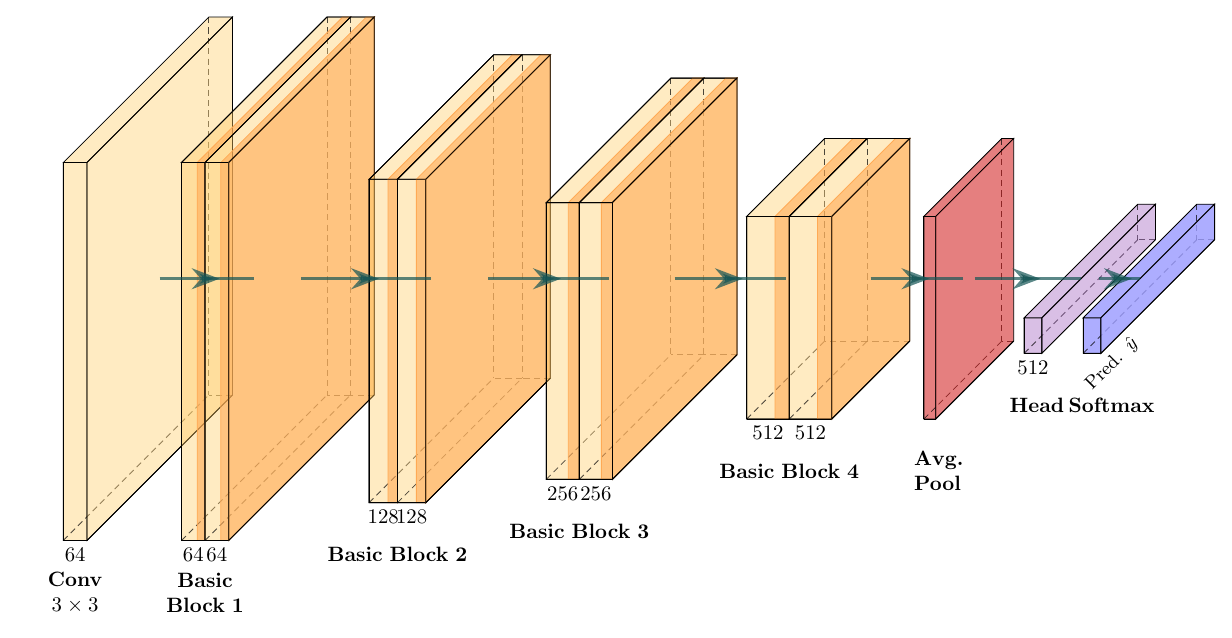}
    \caption[Visualization ResNet-18]{Visualization of the structure of a ResNet-18~\cite{resnet}.}
    \label{fig:resnet18_structure}
\end{figure}

The individual layers of a ResNet-18 are listed in \Cref{tab:model_basic_structure} and a structural overview is depicted in \Cref{fig:resnet18_structure}. There is no difference in the depth of the network or the type of layers between the ResNet-18 and the Slim-ResNet-18. The only difference is the number of base filters $C_b$ for the convolutions in the \emph{Basic Blocks}. The Slim-ResNet-18 uses $C_b\!\!=\!\!20$ while the full ResNet-18 uses $C_b\!\!=\!\!64$. This influences the number of channels for the following operations so that a compression to approximately a tenth of the original size can be achieved.

\begin{table}[t]
    \centering
    \adjustbox{max width=\linewidth}{%
    \begin{tabular}{ccc}
        \toprule
        \multicolumn{3}{c}{\textbf{ResNet-18}} \\ \midrule
        \multicolumn{3}{c}{\textit{$C_b = 20$ for SlimResNet18 and $C_b=64$ for ResNet-18}} \\ \toprule
        \textbf{Layer} & \textbf{Stride} & \textbf{Dimension}  \\ \midrule
        Conv $3\times3$ & 1 & $C_b \times 32 \times 32$ \\ 
        BatchNorm & - & $C_b \times 32 \times 32$ \\ 
        ReLU & - & $C_b \times 32 \times 32$ \\ 
        BasicBlock $C_{in} = C_b$, $C_{out}=C_b$ & 1 & $C_b \times 32 \times 32$ \\ 
        BasicBlock $C_{in} = C_b$, $C_{out}=2 \cdot C_b$ & 2 & $2 \times C_b \times 16 \times 16$ \\ 
        BasicBlock $C_{in} = 2 \cdot C_b$, $C_{out}=3 \cdot C_b$ & 2 & $3 \times C_b \times 8 \times 8$ \\ 
        BasicBlock $C_{in} = 3 \cdot C_b$, $C_{out}=4 \cdot C_b$ & 2 & $4 \times C_b \times 4 \times 4$ \\ 
        AvgPool $4 \times 4$ & 1 & $4 \times C_b \times 1 \times 1$ \\ 
        Linear (Classification Head) & - & $\#\text{classes}$ \\ \bottomrule
    \end{tabular}
    }
    \caption{The network structure is identical for both ResNet-18 and its SlimResNet-18 variant, besides a reduction in the number of base channels~$C_b$.
    }
    \label{tab:model_basic_structure}
\end{table}

\begin{table*}
    \centering
    \adjustbox{max width=\textwidth}{%
    \begin{tabular}{cc}
        \toprule
        \textbf{Method} & \textbf{Hyperparameter} \\ \toprule
        FT & - \\ 
        FZ & freeze after 1st task  \\ 
        Joint &  -  \\ 
        WA~\cite{weigth_align} & $2000$ exemplars, $\tau=2$, $patience=10$ \\ \midrule
        EWC~\cite{ewc} & $\lambda=40000$, $\alpha=0.1$ \\ 
        MAS~\cite{mas} & $\lambda=10$, $\alpha=0.1$  \\ 
        LwF~\cite{lwf} & $\lambda=30$, $\tau=2$ \\ \midrule
        PASS~\cite{pass}& $\tau_{\text{CE}}=0.1$, $\tau_{KD}=2$, $\lambda_{kd} = 10.0$, $\lambda_{aug} = 10.0$  \\ 
        IL2A~\cite{il2a} & $\tau_{\text{CE}}=0.1$, $\lambda_{KD}=10.0$, $\lambda_{seman}=10.0$, $\#mixups=4$ \\ 
        PRAKA~\cite{praka} & $\tau_{\text{CE}}=0.1$, $\lambda_{aug}=15.0$, $\lambda_{KD}=15.0$ \\ 
        SSRE~\cite{ssre} &  $\tau_{\text{CE}}=0.1$, $\lambda_{aug}=10.0$, $\lambda_{KD}=1.0$\\ 
        FeTrIL~\cite{fetril} & AugMix~\cite{augmix} pre-train, fc head, 1-cosine translation \\
        Horde (ours) & original features estimation, {CE} \& {ML} Head, self-supervision, 1 Resnet18, 9 Slim Resnet18s \\ \bottomrule
    \end{tabular}}
    \caption{Overview of approach-specific hyperparameters}
    \label{tab:evaluation_hyperparameters}
\end{table*}

\section{Method Hyperparameters}
\label{sec:hyperparameters}
The hyperparameters for all compared methods are listed in \Cref{tab:evaluation_hyperparameters}. For EWC, MAS and LWF, we perform a gridsearch over their main hyperparameters on the CIL 50/10 scenario, and the one achieving the highest average accuracy are fixed for the repetition tasks. The remaining hyperparameters are the ones recommended by their original authors for the corresponding scenario.

\section{Feature Extractor Training Components}
\label{sec:abl_fe_training}
\Cref{tab:abl_tab_fe_training} provides an overview of the effects of each component in the Feature Extractor and its effect on the average accuracy in the CIL scenario (a). The results have been averaged over 5 seeds. Both the self-supervision from PASS~\cite{pass} as well as the training with the metric learning head are beneficial based on the overall average accuracy. The metric learning head alone without a cross-entropy head is however insufficient for the training of a feature extractor.

\begin{table}[t]
    \centering
    \adjustbox{max width=\linewidth}{%
    \begin{tabular}{cccr}
        \toprule
        \textbf{CE-Head} & \textbf{ML-Head} & \textbf{Self-Supervision~\cite{pass}} & \textbf{Avg. Acc $\uparrow$}  \\ \toprule
        \cmark & \xmark & \xmark & \spacerbullet{$56.91 \pm 0.88$} \\ 
        \xmark & \cmark & \xmark & \spacerbullet{$7.72 \pm 0.85$} \\ 
        \cmark & \cmark & \xmark & \third{$58.68 \pm 0.79$} \\ 
        \cmark & \xmark & \cmark & \second{$60.62 \pm 1.21$} \\ 
        \xmark & \cmark & \cmark & \spacerbullet{$9.76 \pm 1.15$} \\ 
        \cmark & \cmark & \cmark & \best{$63.09 \pm 1.19$} \\ \bottomrule
    \end{tabular}}
    \caption{Ablation study results on different variations of FE training. \best{1st}, \second{2nd} and \third{3rd} best metrics are marked accordingly.}
    \label{tab:abl_tab_fe_training}
\end{table}

\section{Scenario Visualization}
\label{sec:scenario_viz}
\begin{figure}[t]
    \centering
    \includegraphics[width=0.9\linewidth, trim=18 18 18 65, clip]{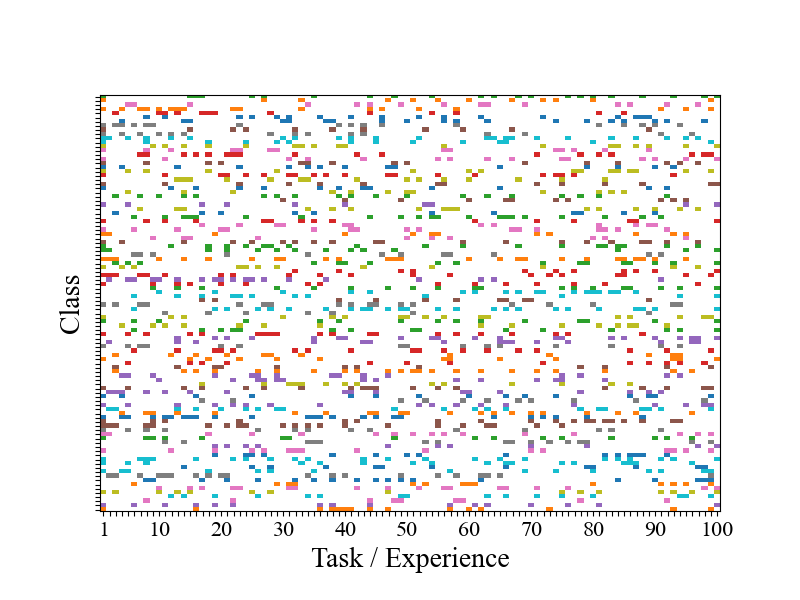}
    \caption{Class distribution visualization of scenario (b), with uniform class occurrence frequency. Each colored block indicates that the class is sampled in the corresponding task.}
    \label{fig:dist_base}
\end{figure}

\begin{figure}[t]
    \centering
    \includegraphics[width=0.9\linewidth, trim=18 18 18 50, clip]{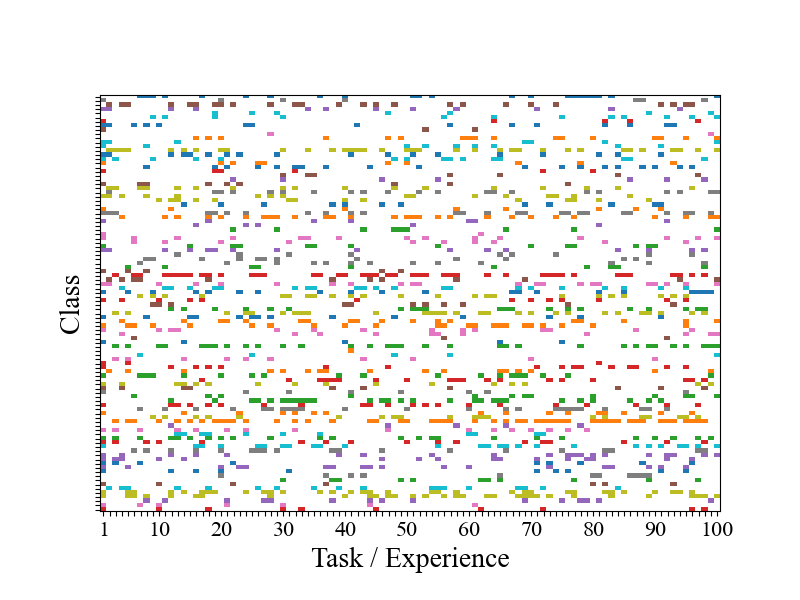}
    \caption{Class distribution visualization of scenario (c), with biased (beta) class occurrence frequency. Each colored block indicates that the class is sampled in the corresponding task.}
    \label{fig:dist_base_2}
\end{figure}

In the proposed experiments we differentiate between a fairly balanced repetition scenario and a biased scenario. The difference between the two repetition frequencies is visualized in \cref{fig:dist_base} and \cref{fig:dist_base_2}. On average both scenarios have 15 classes in each incremental task.

\section{Longer Task Sequence}
\begin{figure*}[t]
    \centering
    \includegraphics[width=0.8\linewidth]{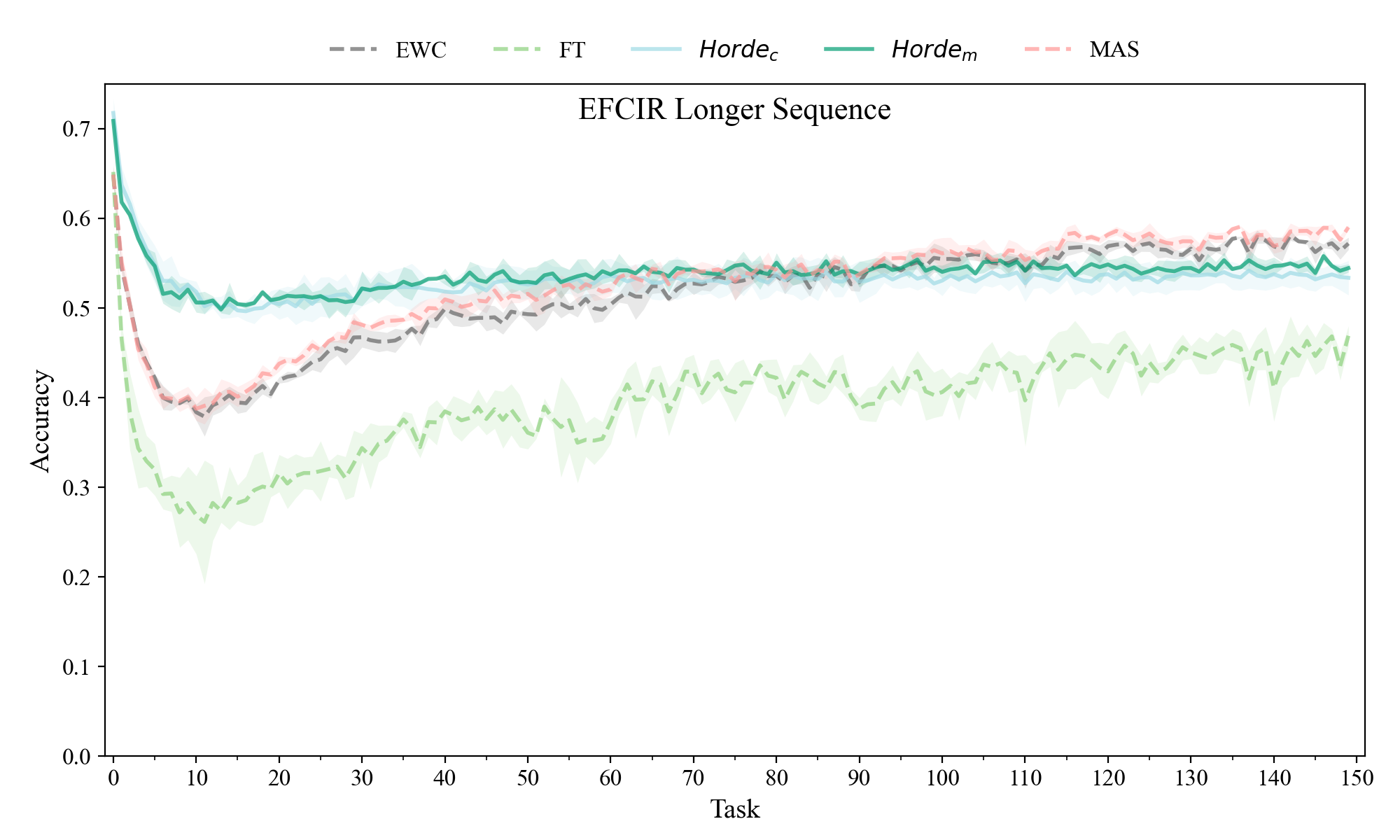}
    \caption{Average Accuracy over an even longer repetition scenario to analyse the trends between different methods. The performance increase of weight-regularization techniques continue }
    \label{fig:longer_task_sequence}
\end{figure*}

The results from scenario (b) indicate a strong accuracy recovery/trend for weight regularization techniques. We further evaluate with even longer task sequences where the number of incremental tasks is increased from 99 to 149. The accuracy on later tasks is very strong on weight-regularization techniques as the overall accuracy trend continues. However, it is important to note that, already in the 100 task scenario, all available training data is used in the task sequence at least once, thus further tasks can only repeat samples and no longer provide any new/incremental training data. Although EWC and MAS both achieve a significant higher final accuracy in the longer task sequence, they are still slightly worse in terms of average accuracy across the whole sequence, since they are less stable in the initial tasks of the sequence. The compared average accuracies for the 100 and 150 task scenarios, as well as final test accuracy after the task sequence, are listed in \Cref{tab:res_longer_sequence}. Furthermore, the accuracy progression is visualized in \Cref{fig:longer_task_sequence}.

\begin{table*}[t]
    \centering
    \begin{tabular}{ccccc}
        \toprule
       \textbf{Method}  & \textbf{Avg. $A_{\text{100}}$} & \textbf{Avg. $A_{\text{150}}$} & \textbf{final $A_{\text{100}}$} & \textbf{final $A_{\text{150}}$} \\ \toprule
        FT & \spacerbullet{$36.2 \pm 2.1$} & \spacerbullet{$39.3 \pm 2.1$} & \spacerbullet{$42.3 \pm 2.7$} & \spacerbullet{$46.9 \pm 1.1$} \\
        EWC & \spacerbullet{$47.7 \pm 3.2$} & \spacerbullet{$51.4 \pm 0.9$} & \third{$54.4 \pm 2.5$} & \second{$57.2 \pm 0.7$}\\
        MAS & \third{$49.3 \pm 2.6$} & \third{$52.5 \pm 0.8$} & \best{$55.6 \pm 2.2$} & \best{$59.0 \pm 0.3$}\\ \midrule
        $Horde_c$ & \best{$54.4 \pm 0.7$} & \second{$53.2 \pm 1.6$} & \second{$55.1 \pm 0.7$} & \third{$54.4 \pm 0.4$} \\
        $Horde_m$ & \second{$53.4 \pm 0.7$} & \best{$53.8 \pm 0.9$} & \spacerbullet{$54.0 \pm 1.1$} & \spacerbullet{$53.4 \pm 1.9$} \\ \bottomrule
    \end{tabular}
    \caption{Comparison between unbiased repetition scenarios of 100 and 150 tasks. While our proposed method is more stable, especially in the initial phases of training. The trend of weight regularization methods continues and the final accuracy continues to increase.}
    \label{tab:res_longer_sequence}
\end{table*}

\section{Scenario Results}
\label{sec:scenario_results}
\begin{figure*}[p]
    \centering
    \includegraphics[width=0.7\linewidth]{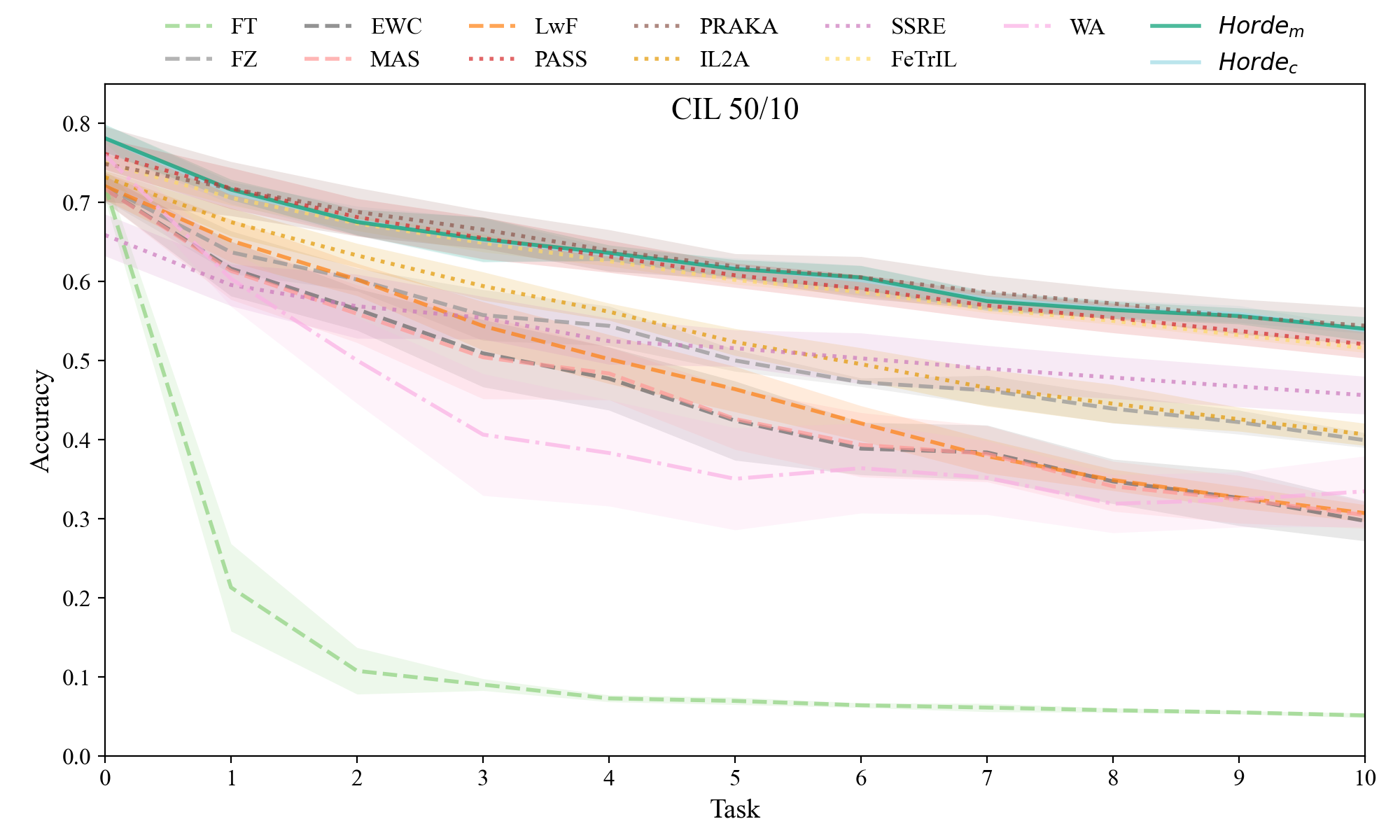}
    \caption{Accuracy development over the task sequence of scenario (a).}
    \label{fig:cil_base_case}
\end{figure*}

\begin{figure*}[p]
    \centering
    \includegraphics[width=0.7\linewidth]{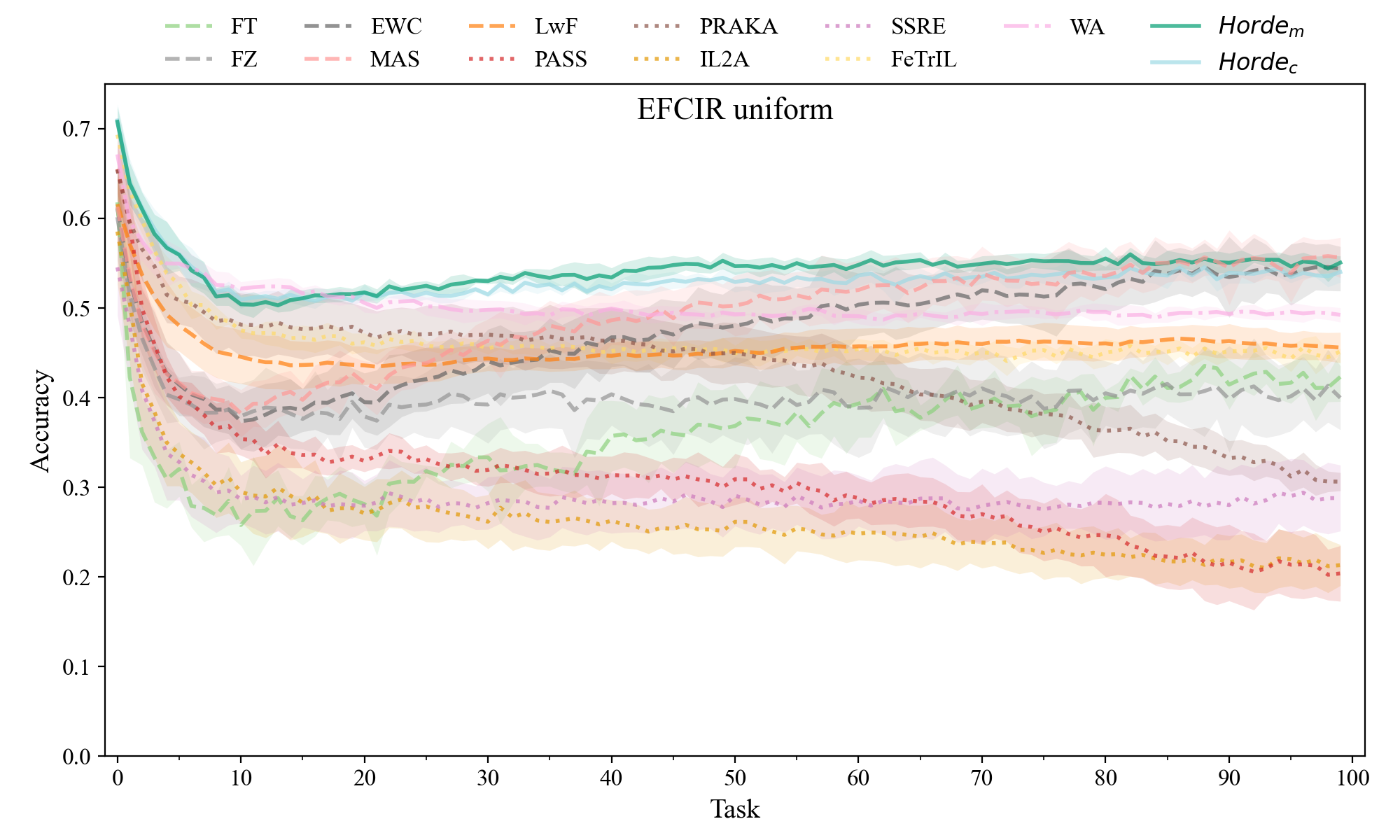}
    \caption{Accuracy development over the task sequence of scenario (b).}
    \label{fig:efcir_supplementary}
\end{figure*}

\begin{figure*}[p]
    \centering
    \includegraphics[width=0.7\linewidth]{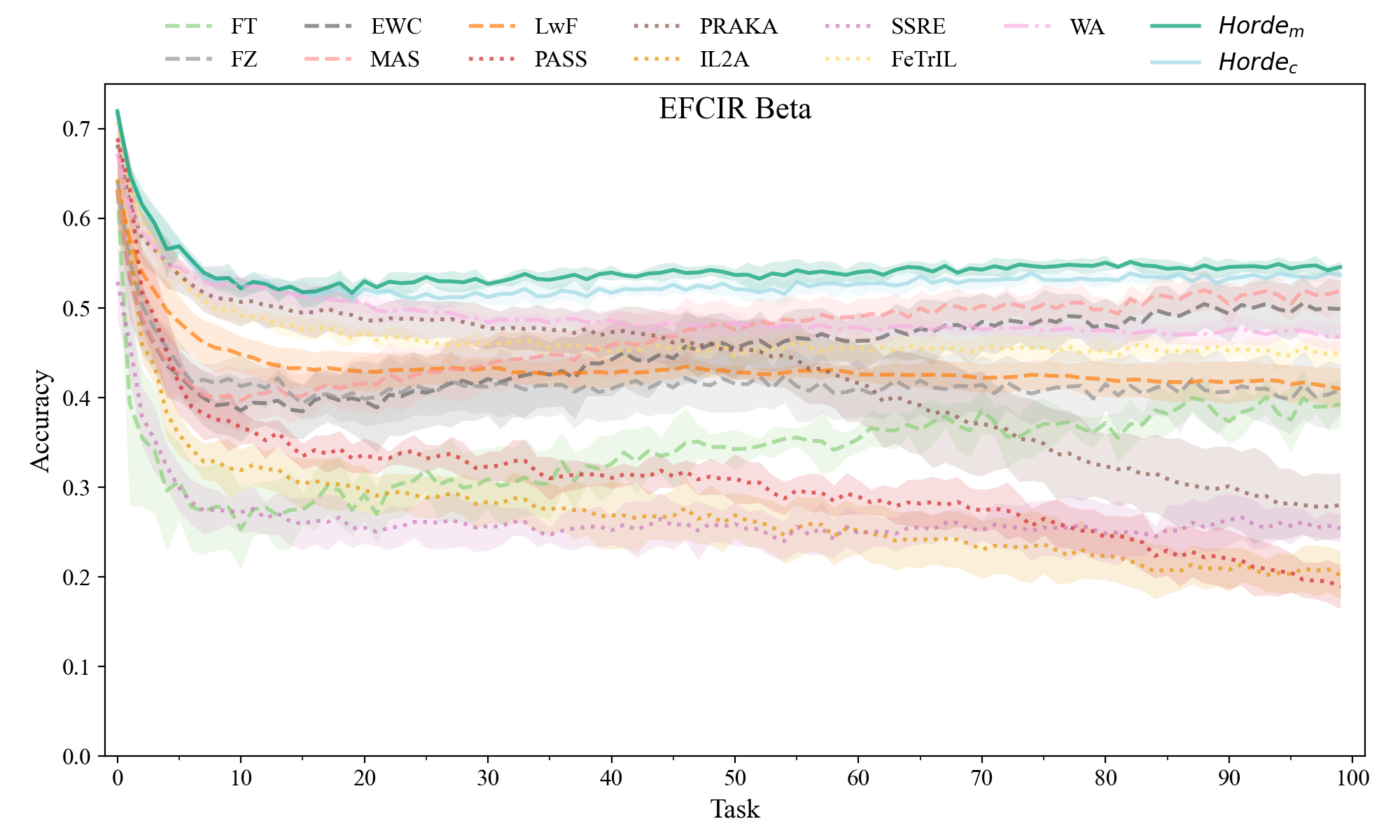}
    \caption{Accuracy development over the task sequence of scenario (c).}
    \label{fig:efcir_beta_supplementary}
\end{figure*}

The following figures visualize the detailed Average Accuracy development over the incremental task sequence. For each method the mean and one standard deviation have been plotted. The results for the class-incremental scenario (a) are listed in \Cref{tab:cil_base_results} and visualized in \Cref{fig:cil_base_case}. The unbiased repetition results fo scenario (b) can be found in \Cref{tab:efcir_uniform_results} and \Cref{fig:efcir_supplementary}. The results of the biased class-repetition scenario (c) are shown in \Cref{tab:efcir_beta_results} and \Cref{fig:efcir_beta_supplementary}.


\begin{table*}[p]
    \centering
    \adjustbox{max width=\textwidth}{%
\begin{tabular}{cccccccccccccc}
\textbf{Method} & \textbf{$\bm{A_{0}}$} & \textbf{$\bm{A_{1}}$} & \textbf{$\bm{A_{2}}$} & \textbf{$\bm{A_{3}}$} & \textbf{$\bm{A_{4}}$} & \textbf{$\bm{A_{5}}$} & \textbf{$\bm{A_{6}}$} & \textbf{$\bm{A_{7}}$} & \textbf{$\bm{A_{8}}$} & \textbf{$\bm{A_{9}}$} & \textbf{$\bm{A_{10}}$} & \textbf{Avg.  $\bm{A \uparrow}$} & \textbf{Avg. $\bm{f \downarrow}$} \\ \toprule
WA  & 76.0 $\pm$ 1.5 & 60.4 $\pm$ 3.5 & 50.0 $\pm$ 5.4 & 40.7 $\pm$ 7.7 & 38.3 $\pm$ 6.7 & 35.1 $\pm$ 6.5 & 36.4 $\pm$ 5.7 & 35.2 $\pm$ 4.7 & 31.9 $\pm$ 3.7 & 32.4 $\pm$ 3.5 & 33.5 $\pm$ 4.5 & 42.7 $\pm$ 2.3 & 33.3 $\pm$ 1.5 \\ 
\midrule
FT  & 72.1 $\pm$ 1.7 & 21.3 $\pm$ 5.5 & 10.8 $\pm$ 2.9 & 9.0 $\pm$ 0.8 & 7.3 $\pm$ 0.4 & 7.0 $\pm$ 0.5 & 6.4 $\pm$ 0.3 & 6.1 $\pm$ 0.5 & 5.8 $\pm$ 0.3 & 5.5 $\pm$ 0.2 & 5.1 $\pm$ 0.3 & 14.2 $\pm$ 1.0 & 57.9 $\pm$ 1.2 \\ 
FZ  & 72.1 $\pm$ 1.7 & 63.7 $\pm$ 2.7 & 60.3 $\pm$ 1.4 & 55.8 $\pm$ 3.2 & 54.4 $\pm$ 2.5 & 50.0 $\pm$ 1.4 & 47.3 $\pm$ 0.5 & 46.2 $\pm$ 1.9 & 43.9 $\pm$ 1.8 & 42.2 $\pm$ 1.5 & 39.9 $\pm$ 0.9 & 52.4 $\pm$ 1.4 & 19.7 $\pm$ 0.9 \\ 
EWC  & 71.7 $\pm$ 1.6 & 61.6 $\pm$ 3.5 & 56.5 $\pm$ 2.6 & 50.9 $\pm$ 4.2 & 47.7 $\pm$ 4.0 & 42.4 $\pm$ 5.0 & 38.9 $\pm$ 3.3 & 38.4 $\pm$ 3.5 & 34.7 $\pm$ 2.8 & 32.6 $\pm$ 3.5 & 29.7 $\pm$ 2.5 & 45.9 $\pm$ 3.0 & 25.8 $\pm$ 1.5 \\ 
MAS  & 71.7 $\pm$ 1.6 & 61.4 $\pm$ 3.6 & 55.9 $\pm$ 3.6 & 50.4 $\pm$ 5.3 & 48.4 $\pm$ 3.3 & 42.6 $\pm$ 3.8 & 39.4 $\pm$ 4.1 & 38.2 $\pm$ 3.6 & 34.1 $\pm$ 3.2 & 32.5 $\pm$ 3.0 & 30.5 $\pm$ 1.7 & 45.9 $\pm$ 2.9 & 25.8 $\pm$ 1.4 \\ 
LwF  & 72.1 $\pm$ 1.7 & 65.2 $\pm$ 2.7 & 60.3 $\pm$ 2.0 & 54.4 $\pm$ 3.1 & 50.2 $\pm$ 3.1 & 46.4 $\pm$ 2.9 & 42.1 $\pm$ 2.2 & 37.9 $\pm$ 2.2 & 34.9 $\pm$ 1.3 & 32.7 $\pm$ 1.4 & 30.7 $\pm$ 1.1 & 47.9 $\pm$ 1.8 & 24.2 $\pm$ 0.8 \\ 
\midrule
PASS  & \third{76.2 $\pm$ 2.0} & \best{71.8 $\pm$ 2.6} & \second{68.2 $\pm$ 2.3} & \second{65.5 $\pm$ 2.6} & 63.2 $\pm$ 2.0 & 60.8 $\pm$ 1.5 & 59.1 $\pm$ 1.8 & 57.0 $\pm$ 1.7 & 55.4 $\pm$ 1.9 & 53.8 $\pm$ 1.7 & 52.1 $\pm$ 1.7 & 62.1 $\pm$ 1.9 & 14.1 $\pm$ 0.5 \\ 
PRAKA  & 74.9 $\pm$ 4.7 & \second{71.8 $\pm$ 3.4} & \best{68.8 $\pm$ 3.1} & \best{66.6 $\pm$ 2.4} & \best{63.9 $\pm$ 2.6} & \best{61.9 $\pm$ 1.6} & \best{60.5 $\pm$ 2.6} & \best{58.7 $\pm$ 2.1} & \best{57.2 $\pm$ 1.9} & \third{55.6 $\pm$ 2.2} & \best{54.4 $\pm$ 2.4} & \best{63.1 $\pm$ 2.6} & \best{11.8 $\pm$ 2.3} \\ 
IL2A  & 73.2 $\pm$ 0.8 & 67.5 $\pm$ 1.8 & 63.3 $\pm$ 1.5 & 59.4 $\pm$ 1.8 & 56.2 $\pm$ 1.1 & 52.4 $\pm$ 1.7 & 49.6 $\pm$ 2.1 & 46.6 $\pm$ 2.3 & 44.6 $\pm$ 2.4 & 42.6 $\pm$ 1.5 & 40.7 $\pm$ 1.4 & 54.2 $\pm$ 1.4 & 19.0 $\pm$ 1.3 \\ 
SSRE  & 65.9 $\pm$ 2.6 & 59.6 $\pm$ 2.6 & 56.9 $\pm$ 4.0 & 55.4 $\pm$ 2.7 & 52.5 $\pm$ 2.9 & 51.6 $\pm$ 2.3 & 50.3 $\pm$ 3.2 & 49.0 $\pm$ 2.9 & 47.9 $\pm$ 2.7 & 46.7 $\pm$ 2.6 & 45.6 $\pm$ 2.4 & 52.9 $\pm$ 2.7 & \second{13.0 $\pm$ 0.8} \\ 
FeTrIL  & 75.0 $\pm$ 1.2 & 70.6 $\pm$ 0.9 & 67.4 $\pm$ 0.9 & 64.9 $\pm$ 1.1 & 62.6 $\pm$ 0.6 & 60.2 $\pm$ 0.5 & 58.7 $\pm$ 0.5 & 56.5 $\pm$ 0.5 & 55.0 $\pm$ 0.5 & 53.2 $\pm$ 0.4 & 51.6 $\pm$ 0.6 & 61.4 $\pm$ 0.4 & \third{13.6 $\pm$ 0.8} \\ 
\midrule
$Horde_{m}$  & \second{78.1 $\pm$ 1.7} & \third{71.6 $\pm$ 1.3} & 67.5 $\pm$ 1.6 & \third{65.3 $\pm$ 2.8} & \second{63.6 $\pm$ 1.0} & \third{61.6 $\pm$ 1.2} & \second{60.5 $\pm$ 1.5} & \third{57.5 $\pm$ 1.2} & \third{56.4 $\pm$ 1.1} & \second{55.7 $\pm$ 1.0} & \second{54.0 $\pm$ 1.5} & \second{62.9 $\pm$ 1.2} & 15.2 $\pm$ 0.7 \\ 
$Horde_{c}$  & \best{78.2 $\pm$ 1.7} & 71.2 $\pm$ 1.4 & \third{67.7 $\pm$ 1.9} & 65.2 $\pm$ 2.8 & \third{63.4 $\pm$ 0.8} & \second{61.8 $\pm$ 1.2} & \third{60.2 $\pm$ 1.9} & \second{57.9 $\pm$ 1.0} & \second{56.6 $\pm$ 1.1} & \best{55.9 $\pm$ 1.1} & \third{53.8 $\pm$ 0.4} & \third{62.9 $\pm$ 1.2} & 15.3 $\pm$ 0.7 \\ 
\bottomrule
\end{tabular}
    }
    \caption{Results for the baseline CIL 50/10 scenario (a). \best{1st}, \second{2nd} and \third{3rd} best metrics are marked accordingly.}
    \label{tab:cil_base_results}
\end{table*}

\begin{table*}[p]
    \centering
    \adjustbox{max width=\textwidth}{%
\begin{tabular}{cccccccccc}
\textbf{Method} & \textbf{$\bm{A_{0}}$} & \textbf{$\bm{A_{10}}$} & \textbf{$\bm{A_{20}}$} & \textbf{$\bm{A_{40}}$} & \textbf{$\bm{A_{60}}$} & \textbf{$\bm{A_{80}}$} & \textbf{$\bm{A_{99}}$} & \textbf{Avg.  $\bm{A \uparrow}$} & \textbf{Avg. $\bm{f \downarrow}$} \\ \toprule
WA  & 67.1 $\pm$ 2.4 & 52.2 $\pm$ 0.8 & 50.6 $\pm$ 1.1 & 49.9 $\pm$ 0.7 & 49.1 $\pm$ 0.8 & 49.6 $\pm$ 0.9 & 49.2 $\pm$ 0.8 & 50.4 $\pm$ 0.2 & 16.7 $\pm$ 2.4 \\ 
\midrule
FT  & 61.8 $\pm$ 4.3 & 25.8 $\pm$ 2.3 & 28.1 $\pm$ 2.2 & 35.7 $\pm$ 3.9 & 39.3 $\pm$ 3.8 & 40.0 $\pm$ 0.9 & 42.3 $\pm$ 2.7 & 36.2 $\pm$ 2.1 & 25.6 $\pm$ 2.7 \\ 
FZ  & 60.1 $\pm$ 4.8 & 37.9 $\pm$ 3.7 & 37.9 $\pm$ 3.4 & 40.4 $\pm$ 3.5 & 38.9 $\pm$ 5.5 & 40.5 $\pm$ 3.6 & 40.0 $\pm$ 3.6 & 40.2 $\pm$ 4.0 & 20.0 $\pm$ 1.6 \\ 
EWC  & 61.1 $\pm$ 4.1 & 37.4 $\pm$ 3.1 & 39.5 $\pm$ 3.3 & 46.7 $\pm$ 3.3 & 50.4 $\pm$ 3.5 & 52.1 $\pm$ 2.5 & \third{54.4 $\pm$ 2.5} & 47.7 $\pm$ 3.2 & \second{13.5 $\pm$ 1.5} \\ 
MAS  & 61.3 $\pm$ 4.1 & 38.2 $\pm$ 2.6 & 41.6 $\pm$ 2.5 & \third{48.6 $\pm$ 2.7} & \third{52.0 $\pm$ 3.3} & \second{53.6 $\pm$ 1.9} & \best{55.6 $\pm$ 2.2} & \third{49.3 $\pm$ 2.6} & \best{12.0 $\pm$ 1.8} \\ 
LwF  & 61.6 $\pm$ 4.2 & 44.5 $\pm$ 3.0 & 43.6 $\pm$ 2.2 & 44.7 $\pm$ 1.5 & 45.7 $\pm$ 2.0 & 46.1 $\pm$ 1.9 & 45.6 $\pm$ 1.7 & 45.7 $\pm$ 1.9 & \third{15.9 $\pm$ 2.8} \\ 
\midrule
PASS  & 65.5 $\pm$ 2.7 & 35.4 $\pm$ 1.8 & 32.9 $\pm$ 1.2 & 31.4 $\pm$ 2.1 & 28.8 $\pm$ 2.2 & 24.6 $\pm$ 4.6 & 20.4 $\pm$ 3.1 & 30.2 $\pm$ 1.9 & 35.3 $\pm$ 2.2 \\ 
PRAKA  & 65.4 $\pm$ 3.3 & \third{48.1 $\pm$ 3.5} & \third{47.2 $\pm$ 2.9} & 46.4 $\pm$ 3.7 & 42.2 $\pm$ 3.0 & 36.3 $\pm$ 2.7 & 30.6 $\pm$ 1.0 & 43.1 $\pm$ 2.1 & 22.3 $\pm$ 2.4 \\ 
IL2A  & 58.5 $\pm$ 5.8 & 29.5 $\pm$ 3.3 & 27.1 $\pm$ 3.0 & 26.3 $\pm$ 2.3 & 24.8 $\pm$ 3.0 & 22.5 $\pm$ 2.6 & 21.3 $\pm$ 2.3 & 26.3 $\pm$ 3.0 & 32.2 $\pm$ 2.9 \\ 
SSRE  & 54.5 $\pm$ 5.3 & 28.7 $\pm$ 2.7 & 27.9 $\pm$ 2.8 & 28.2 $\pm$ 3.4 & 28.5 $\pm$ 2.8 & 28.3 $\pm$ 4.1 & 28.8 $\pm$ 3.7 & 29.2 $\pm$ 3.5 & 25.4 $\pm$ 2.1 \\ 
FeTrIL  & \third{69.3 $\pm$ 1.1} & 47.5 $\pm$ 0.9 & 46.2 $\pm$ 1.0 & 45.2 $\pm$ 1.4 & 45.4 $\pm$ 1.1 & 45.0 $\pm$ 1.5 & 45.2 $\pm$ 0.3 & 46.5 $\pm$ 0.7 & 22.9 $\pm$ 0.7 \\ 
\midrule
$Horde_{m}$  & \second{70.8 $\pm$ 1.7} & \second{50.4 $\pm$ 1.1} & \best{51.7 $\pm$ 1.0} & \best{53.4 $\pm$ 1.1} & \best{54.8 $\pm$ 1.1} & \best{55.5 $\pm$ 1.3} & \second{55.1 $\pm$ 0.7} & \best{54.4 $\pm$ 0.7} & 16.4 $\pm$ 1.5 \\ 
$Horde_{c}$  & \best{70.9 $\pm$ 1.9} & \best{50.9 $\pm$ 1.0} & \second{51.7 $\pm$ 0.9} & \second{52.1 $\pm$ 0.7} & \second{53.6 $\pm$ 1.2} & \third{53.4 $\pm$ 1.4} & 54.0 $\pm$ 1.1 & \second{53.4 $\pm$ 0.7} & 17.6 $\pm$ 1.6 \\ 
\bottomrule
\end{tabular}
    }
    \caption{Results for the EFCIR-U scenario (b). \best{1st}, \second{2nd} and \third{3rd} best metrics are marked accordingly.}
    \label{tab:efcir_uniform_results}
\end{table*}

\begin{table*}[p]
    \centering
    \adjustbox{max width=\textwidth}{%
\begin{tabular}{cccccccccc}
\textbf{Method} & \textbf{$\bm{A_{0}}$} & \textbf{$\bm{A_{10}}$} & \textbf{$\bm{A_{20}}$} & \textbf{$\bm{A_{40}}$} & \textbf{$\bm{A_{60}}$} & \textbf{$\bm{A_{80}}$} & \textbf{$\bm{A_{99}}$} & \textbf{Avg.  $\bm{A \uparrow}$} & \textbf{Avg. $\bm{f \downarrow}$} \\ \toprule
WA  & 67.2 $\pm$ 1.8 & 52.4 $\pm$ 1.2 & 50.6 $\pm$ 1.1 & 48.6 $\pm$ 1.1 & 47.9 $\pm$ 1.6 & 47.5 $\pm$ 0.8 & 46.8 $\pm$ 1.6 & 49.2 $\pm$ 0.7 & 18.0 $\pm$ 1.6 \\ 
\midrule
FT  & 63.2 $\pm$ 4.3 & 25.3 $\pm$ 4.5 & 29.2 $\pm$ 3.0 & 32.6 $\pm$ 3.2 & 35.4 $\pm$ 1.5 & 37.2 $\pm$ 0.9 & 39.3 $\pm$ 2.7 & 34.2 $\pm$ 2.0 & 29.0 $\pm$ 2.6 \\ 
FZ  & 64.2 $\pm$ 4.3 & 41.3 $\pm$ 3.4 & 39.9 $\pm$ 3.0 & 41.0 $\pm$ 3.5 & 41.0 $\pm$ 3.2 & 41.5 $\pm$ 2.9 & 40.9 $\pm$ 2.8 & 41.7 $\pm$ 3.1 & 22.5 $\pm$ 1.9 \\ 
EWC  & 63.2 $\pm$ 4.3 & 38.6 $\pm$ 3.5 & 39.6 $\pm$ 4.6 & 44.2 $\pm$ 4.1 & 46.3 $\pm$ 3.2 & 48.1 $\pm$ 3.0 & 49.9 $\pm$ 3.2 & 45.5 $\pm$ 3.2 & \third{17.8 $\pm$ 1.8} \\ 
MAS  & 63.2 $\pm$ 4.3 & 39.7 $\pm$ 3.5 & 41.6 $\pm$ 3.1 & 45.9 $\pm$ 3.7 & \third{49.0 $\pm$ 2.1} & \third{49.9 $\pm$ 1.2} & \third{51.9 $\pm$ 2.0} & \third{47.1 $\pm$ 2.3} & \best{16.1 $\pm$ 2.1} \\ 
LwF  & 63.2 $\pm$ 4.3 & 44.8 $\pm$ 2.0 & 42.9 $\pm$ 2.1 & 42.7 $\pm$ 1.6 & 42.6 $\pm$ 0.7 & 42.0 $\pm$ 1.9 & 41.0 $\pm$ 2.2 & 43.5 $\pm$ 0.8 & 19.8 $\pm$ 4.1 \\ 
\midrule
PASS  & 68.9 $\pm$ 2.0 & 36.7 $\pm$ 1.9 & 33.6 $\pm$ 1.9 & 31.0 $\pm$ 1.2 & 28.9 $\pm$ 1.9 & 24.4 $\pm$ 2.4 & 18.9 $\pm$ 2.4 & 30.6 $\pm$ 1.4 & 38.3 $\pm$ 1.6 \\ 
PRAKA  & 68.2 $\pm$ 2.2 & \third{50.7 $\pm$ 2.7} & \third{48.6 $\pm$ 1.7} & \third{47.3 $\pm$ 2.5} & 41.6 $\pm$ 4.3 & 32.3 $\pm$ 3.7 & 28.0 $\pm$ 3.6 & 42.6 $\pm$ 3.2 & 25.6 $\pm$ 1.9 \\ 
IL2A  & 64.4 $\pm$ 4.1 & 31.9 $\pm$ 2.4 & 29.7 $\pm$ 2.6 & 26.9 $\pm$ 3.8 & 25.2 $\pm$ 2.6 & 22.3 $\pm$ 2.6 & 20.2 $\pm$ 2.7 & 27.2 $\pm$ 2.5 & 37.2 $\pm$ 1.7 \\ 
SSRE  & 52.9 $\pm$ 4.1 & 27.2 $\pm$ 2.4 & 25.5 $\pm$ 1.7 & 25.2 $\pm$ 1.6 & 24.9 $\pm$ 2.3 & 24.9 $\pm$ 3.2 & 25.4 $\pm$ 1.5 & 26.5 $\pm$ 2.2 & 26.4 $\pm$ 2.1 \\ 
FeTrIL  & \third{70.7 $\pm$ 1.5} & 49.1 $\pm$ 0.9 & 47.4 $\pm$ 0.5 & 45.1 $\pm$ 1.6 & 45.5 $\pm$ 1.8 & 45.2 $\pm$ 1.6 & 45.0 $\pm$ 1.5 & 46.9 $\pm$ 0.9 & 23.8 $\pm$ 1.2 \\ 
\midrule
$Horde_{m}$  & \best{72.0 $\pm$ 0.8} & \second{52.2 $\pm$ 1.1} & \best{53.0 $\pm$ 0.6} & \best{54.0 $\pm$ 0.9} & \best{54.0 $\pm$ 1.2} & \best{55.1 $\pm$ 0.8} & \best{54.6 $\pm$ 0.7} & \best{54.3 $\pm$ 0.4} & \second{17.7 $\pm$ 1.0} \\ 
$Horde_{c}$  & \second{71.7 $\pm$ 0.8} & \best{52.5 $\pm$ 0.7} & \second{52.1 $\pm$ 0.7} & \second{51.7 $\pm$ 0.7} & \second{52.7 $\pm$ 1.4} & \second{53.2 $\pm$ 0.8} & \second{53.7 $\pm$ 0.4} & \second{53.1 $\pm$ 0.4} & 18.5 $\pm$ 1.1 \\ 
\bottomrule
\end{tabular}
    }
    \caption{Results for the EFCIR-B scenario (c). \best{1st}, \second{2nd} and \third{3rd} best metrics are marked accordingly.}
    \label{tab:efcir_beta_results}
\end{table*}

\end{document}